\documentclass[journal]{IEEEtran}
\usepackage{amsmath,amsfonts}
\usepackage{algorithm}
\usepackage{algpseudocode}
\usepackage{array}
\usepackage{subcaption}
\usepackage{xspace}

\newcommand{\eg}{e.g.\xspace}
\DeclareRobustCommand{\ORF}[1][\large]{%
  \kern0.02em%
  {\scriptsize ORF}%
  \kern0.02em%
}

\usepackage{textcomp}
\usepackage{stfloats}
\usepackage{url}
\usepackage{verbatim}
\usepackage{graphicx}
\usepackage{cite}
\usepackage{hyperref}
\usepackage{amssymb}
\usepackage{colortbl}
\usepackage{multirow}
\usepackage{array}
\usepackage{booktabs}
\usepackage[dvipsnames]{xcolor}
\definecolor{LightCyan}{rgb}{0.88,1,1}
\usepackage{enumitem} 
\usepackage{csquotes}
\usepackage{cleveref}
\Crefname{figure}{Fig.}{Figs.}
\crefname{figure}{Fig.}{Figs.}
\usepackage{marvosym}
\usepackage{placeins}

\begin{document}

\title{Cross-Sample Relational Fusion: \\Unifying Domain Generalization and Class-Incremental Learning}

\author{Zhen-Hao Xie, Yan Wang, Hao Sun, Han-Jia Ye, De-Chuan Zhan, Da-Wei Zhou
}

\markboth{Journal of \LaTeX\ Class Files,~Vol.~14, No.~8, August~2021}%
{Shell \MakeLowercase{\textit{et al.}}: A Sample Article Using IEEEtran.cls for IEEE Journals}

\maketitle

\footnotetext{
This work is partially supported by NSFC (62506160), Basic Research Program of Jiangsu (BK 20251251), JSTJ-2025-147, Fundamental and Interdisciplinary Disciplines Breakthrough Plan of the Ministry of Education of China (No. JYB2025XDXM118), the 111 Center (No. B26023). 	(Corresponding author: Da-Wei Zhou.)

Zhen-Hao Xie, Yan Wang, Hao Sun, Han-Jia Ye, De-Chuan Zhan and Da-Wei Zhou are with the School of Artificial Intelligence and the State Key Laboratory of Novel Software Technology, Nanjing University, Nanjing, China (e-mail: wenzh@lamda.nju.edu.cn, wangy@lamda.nju.edu.cn, sunhao@lamda.nju.edu.cn, yehj@lamda.nju.edu.cn, zhandc@nju.edu.cn, zhoudw@lamda.nju.edu.cn).}
\begin{abstract}
Class-Incremental Learning (CIL) requires a learning system to learn new classes while retaining previously learned knowledge. However, in real-world scenarios such as autonomous driving, a system trained on urban roads in sunny weather may later need to operate in rural or highway environments with different traffic patterns and weather conditions. This requires the model not only to overcome catastrophic forgetting, but also to effectively handle domain shifts. In this paper, we propose CrOss-sample Relational Fusion (C\ORF), a unified framework to address domain shift and catastrophic forgetting simultaneously. To enhance generalizability, we perform selective refinement of training samples by leveraging spatial contribution maps to highlight semantically informative regions. Furthermore, we incorporate predictive confidence to adaptively weigh samples, thereby facilitating the learning of domain-agnostic representations. To alleviate forgetting, we propose a cascaded distillation framework that captures cross-sample relational dependencies across multiple feature hierarchies, enabling multi-grained knowledge transfer from previous tasks. C\ORF \ can be seamlessly integrated into existing CIL algorithms to enhance their generalizability, achieving competitive performance across various benchmark datasets. Code is available: \href{https://github.com/LAMDA-CL/TMM26-CORF}{https://github.com/LAMDA-CL/TMM26-CORF}.
\end{abstract}

\section{Introduction}
The advent of deep learning has significantly advanced neural networks, enabling strong performance across real-world applications~\cite{DBLP:journals/nature/LeCunBH15},~\cite{DBLP:journals/ijon/GuoLOLWL16}. In open-world scenarios, data often arrives continuously, requiring models to incrementally incorporate new class knowledge. This process is known as Class-Incremental Learning (CIL)~\cite{DBLP:conf/wacv/ZhangZGLTHZK20, DBLP:journals/pami/ZhouWQYZL24}, in which models must continuously acquire new class knowledge while retaining previously learned knowledge. A central challenge in CIL is catastrophic forgetting~\cite{kirkpatrick2017overcoming}, as the model tends to overwrite old knowledge when learning new tasks. However, even when forgetting is alleviated, in practical scenarios the model may still encounter data from domains that differ from the training distribution, resulting in domain shifts~\cite{DBLP:conf/cvpr/Luo0GYY19, DBLP:conf/ijcai/0001LLOQ21} that further intensify the challenge. For example, in autonomous driving or robotics, new object categories such as unfamiliar road signs, emerging transport modes, and diverse environments frequently appear, demanding continual adaptation without erasing previously acquired visual knowledge.

Existing CIL methods~\cite{DBLP:journals/pami/MasanaLTMBW23,DBLP:conf/nips/XuZ18},~\cite{DBLP:conf/nips/ZhuCZL21,DBLP:conf/cvpr/ShiZLJFTBT22} primarily aim to mitigate catastrophic forgetting~\cite{DBLP:journals/tmm/ThuseethanRY22,DBLP:journals/tmm/LiGXWLLWZ24},~\cite{DBLP:conf/eccv/0002ZESZLRSPDP22}. However, they often struggle under domain shifts due to the lack of a generalization mechanism. In contrast, Domain Generalization (DG) techniques~\cite{DBLP:conf/cvpr/LiPWK18},~\cite{DBLP:conf/cvpr/0001GXL21},~\cite{DBLP:conf/iclr/YangW023},~\cite{DBLP:conf/aaai/LiYSH18} enhance robustness across domains by learning domain-invariant representations~\cite{DBLP:journals/tmm/JinLZC22},~\cite{DBLP:journals/tkde/WangLLOQLCZY23,DBLP:journals/tmm/NiuYMXLCTL24}. Nevertheless, they lack a mechanism to resist forgetting and therefore fail to preserve prior knowledge when learning from continuous data streams.

\begin{figure}
    \centering
\includegraphics[width=\linewidth]{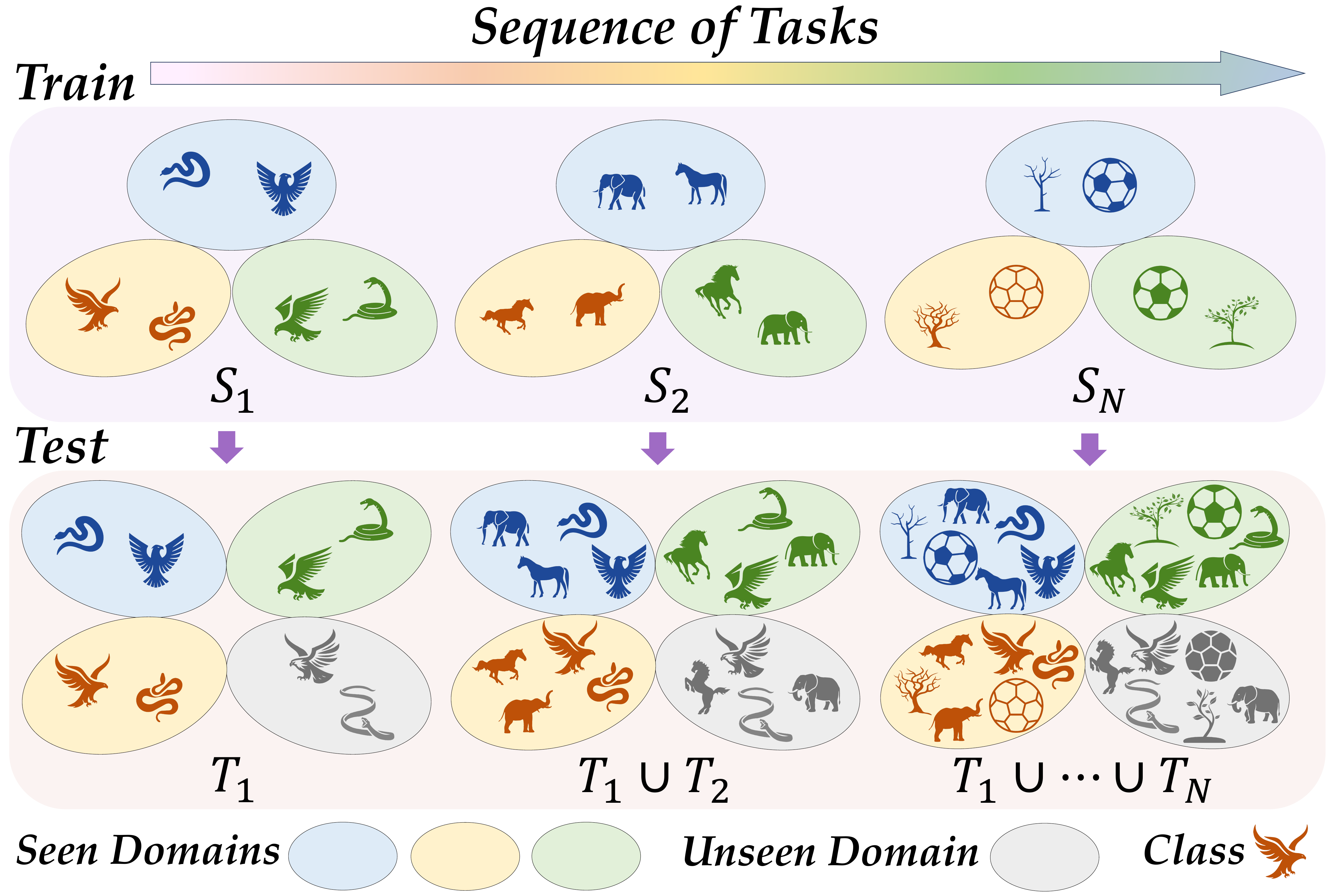}
    \caption{Illustration of CDCIL. During training, novel classes are introduced incrementally, with samples drawn from multiple seen domains. During inference, the model is evaluated on all previously seen classes across both seen and unseen domains. For example, in task 2, although only horse and elephant are observed during training (from seen domains shown in blue, yellow, and green), inference covers all encountered classes (\textit{e.g.}, snake, eagle, horse, and elephant) across both seen and unseen domains (shown in grey).}
    \label{fig:setting}
\end{figure}

Both CIL and DG fail to address the challenge of learning new classes sequentially under domain shift, leading to poor generalizability or severe forgetting. To address both forgetting and domain shift in a unified manner, we formulate the problem as Cross-Domain Class-Incremental Learning (CDCIL). As illustrated in Fig.~\ref{fig:setting}, the model must learn new classes incrementally while generalizing across domains.

Tackling CDCIL presents two key challenges: \textbf{1)} Constructing a domain-agnostic representation space. Since the model is evaluated on an unseen domain, it must effectively handle domain shifts by learning representations that are compact within the same class across domains while remaining well-separated between classes. \textbf{2)} Developing a domain-agnostic strategy to mitigate forgetting. Since samples come from diverse domains, mitigating forgetting requires a fine-grained approach that captures both inter-class and cross-domain dependencies. Instead of supervising each sample independently, the model should retain knowledge across structurally related samples to preserve shared patterns and interactions.

In this paper, we propose CrOss-sample Relational Fusion (C\ORF) to address the challenges of CDCIL. To enhance cross-domain generalizability, we introduce a contribution- and confidence-sensitive refinement mechanism. High-confidence samples are blended with same-class counterparts from different domains in low-contribution regions, while low-confidence samples are fused with different-class instances from the same domain in high-contribution regions. Jointly optimizing original and synthetic samples not only enhances representation discrimination but also facilitates the construction of a domain-agnostic representation space. We further propose a hierarchical kernel-based distillation strategy that aligns globalized information and embeds cross-domain and inter-class relational cues into the transfer process. Together, the refinement and distillation components complement each other by enhancing both the discriminative power and structural consistency of learned representations, thereby promoting robust and generalizable feature learning in CDCIL.

\section{Related Works}
\paragraph{Class-Incremental Learning} To address the challenge of catastrophic forgetting~\cite{DBLP:journals/neco/FrenchC02, kirkpatrick2017overcoming, DBLP:conf/iclr/RamaseshLD22} in CIL~\cite{DBLP:journals/pami/ZhouWQYZL24, DBLP:conf/cvpr/Wen0DQW0024},~\cite{DBLP:conf/nips/Li00LH24, DBLP:journals/tmm/DuLLHFXXC24}, incremental learning methods must maintain the performance of previously learned tasks while adapting to new ones. CIL algorithms can be categorized into several categories. Specifically, replay-based methods~\cite{DBLP:journals/tmm/LiGXWLLWZ24, petit2023fetril} focus on revisiting old knowledge to avoid forgetting, using an exemplar set to store representative instances from previous tasks and reintroduce them during the learning of new tasks. Apart from directly saving instances in the exemplar set, generative models are also widely applied to model the distribution of previous tasks for replay, \textit{\eg}, GAN~\cite{DBLP:conf/cvpr/WangYLHL021}, diffusion model~\cite{DBLP:conf/icml/GaoL23a} and VAE~\cite{DBLP:conf/cvpr/JiangCC21}. Data regularization-based methods~\cite{DBLP:conf/iclr/RiemerCALRTT19},~\cite{aljundi2019gradient,DBLP:journals/tmm/YangFXRDTAR23} mitigate catastrophic forgetting by imposing additional constraints on the loss function. These regularization methods can be applied to the weights by estimating the importance of the model parameters, ensuring that relevant weights do not drift significantly. Dynamic network-based methods~\cite{DBLP:conf/iclr/YoonYLH18},~\cite{DBLP:conf/cvpr/SmithKGCKAPFK23,DBLP:conf/iclr/0001WYZ23}  enhance the model's representation capacity by dynamically adjusting and expanding the network in incremental learning, thereby better adapting to new categories. Such methods are typically categorized into three types of expansions: neuron expansion~\cite{DBLP:conf/iclr/YoonYLH18,DBLP:conf/nips/XuZ18,9676463}, backbone expansion~\cite{DBLP:conf/iclr/0001WYZ23, DBLP:conf/cvpr/LiuSS21}, and prompt expansion~\cite{DBLP:conf/cvpr/SmithKGCKAPFK23,DBLP:conf/cvpr/0002ZL0SRSPDP22}. Model rectification-based methods~\cite{DBLP:conf/cvpr/WuCWYLGF19,DBLP:conf/iccv/BelouadahP19} mitigate catastrophic forgetting by correcting weights, logits, or embeddings, which enables a more balanced classification between old and new classes. More recently, CMoA~\cite{10891550} introduces a contrastive mixture-of-adapters architecture for generalized few-shot continual learning, dynamically allocating adapter modules within a transformer backbone and employing contrastive objectives to improve representation robustness and mitigate forgetting under limited data. 
From a complementary perspective, recent studies have also advanced the theoretical and optimization foundations of CIL. 
For instance, \cite{lin2023theory} analyzes the factors influencing forgetting and generalization, BCL~\cite{raghavan2021formalizing} provides a formal treatment of the stability–plasticity dilemma, and PGM~\cite{wan2025probabilistic} proposes a probabilistic masking and discrete optimization framework to enhance robustness in CIL.

\paragraph{Domain Generalization} DG~\cite{zhou2022domain}, \cite{DBLP:conf/cvpr/PengZ024a}, \cite{DBLP:conf/nips/DayalBCM0B23}, \cite{DBLP:journals/tmm/LiuXLTZ23}, \cite{DBLP:journals/tmm/LiLLG23} focuses on training models that can generalize to previously unseen domains without access to out-of-distribution target domain data during training. One notable direction involves causal inference~\cite{DBLP:conf/cvpr/0016YC022, DBLP:conf/cvpr/LvLLZLWL22}, which aims to disentangle class-relevant and domain-independent features to improve robustness against domain shifts.  The second line of research focuses on modifying the learning objective~\cite{DBLP:conf/wacv/BucciBCT22, DBLP:journals/corr/abs-2007-02931}, aiming to improve the model’s ability to adapt across different domains and tasks by altering its optimization objective. I\textsuperscript{3}C  \cite{DBLP:journals/tmm/ZhouWD25} aims to learn intrinsic intra-class invariance for domain generalization. It achieves this by progressively aligning each sample with its class-specific domain-invariant prototype through an adversarial perturbation-based objective. The third line of research centers on prompt-based methods~\cite{DBLP:journals/corr/abs-2209-14926, DBLP:conf/iccv/ChoNKYK23}, which leverage learnable or dynamically generated prompts to guide feature learning. To produce domain-invariant representations and enhance generalization across unseen domains, this method is often combined with contrastive objectives. DPStyler \cite{DBLP:journals/tmm/TangWQG25} introduces a dynamic prompt-styling mechanism for source-free domain generalization, generating domain-invariant representations via dynamic input modifications and contrastive learning. HCVP \cite{DBLP:journals/tmm/ZhouHCHZLYZ25} proposes a hierarchical contrastive visual prompt framework that generates domain- and task-aware prompts and employs contrastive learning to enhance domain-invariant representations. Additionally, the last line of research uses domain augmentation strategies to increase the diversity of the training data by introducing synthetic variations in domain conditions. These strategies typically include three categories. Domain randomization~\cite{DBLP:conf/eccv/NazariK20, DBLP:conf/cvpr/0001GXL21}, which generates synthetic data by randomizing domain-specific parameters such as lighting, textures, or object positions to improve generalization. Adversarial data augmentation~\cite{DBLP:journals/tmm/DingGZWF24, DBLP:conf/nips/YangCSW21}, which uses adversarial techniques to generate perturbations in the data, making the model more robust to changes in the input domain. FedDG~\cite{DBLP:journals/tmm/XuZZWW24} combines federated learning and adversarial training to generate privacy-preserving adversarial samples that simulate the domain variation. Generative methods~\cite{zhou2021mixstyle} use generative models such as GANs to synthesize novel domain variations, thereby expanding the variety of training examples.

\paragraph{Cross-Domain Class-Incremental Learning}
There are also studies focusing on domain generalization in the context of continual learning. CDCIL addresses both catastrophic forgetting caused by semantic shifts and domain generalization beyond the training domains. Compared to CIL and DG, CDCIL is more challenging and less explored. MSL+MOV~\cite{DBLP:conf/cvpr/SimonFT0SSHC22} is the first method to address these challenges. It suggests learning Mahalanobis similarity metrics to build robust classifiers, while considering the data's covariance structure to better distinguish between old and new classes. Additionally, it introduces an exponential moving average framework for knowledge distillation, enabling gradual knowledge transfer from old to new models and reducing catastrophic forgetting. TRIPS~\cite{peng2024multivariate}  is another related work that proposes an exemplar-free method based on triplet loss with pseudo old-class feature sampling. This method aims to continuously estimate the drift of the mean and covariance matrices of each class prototype as tasks are updated.

\section{Preliminaries}
In this section, we provide an overview of CDCIL and corresponding baselines in CIL and DG. 
\subsection{CDCIL: Task and Evaluation Protocol}
\paragraph{Cross-Domain Class-Incremental Learning}
Formally, in CDCIL, a model is trained continually on a sequence of tasks~\cite{DBLP:conf/cvpr/RebuffiKSL17} to build a unified classifier. The training data is provided as a stream, represented by $\{\mathcal{S}_1, \mathcal{S}_2, \dots, \mathcal{S}_N\}$, where $\mathcal{S}_n = \{\mathcal{D}^1_n, \mathcal{D}^2_n, \dots, \mathcal{D}^{K}_n\}$ corresponds to the $n$-th training set, containing data from $K$ seen domains. $\mathcal{D}_n^k=\{(\mathbf{x}_i,y_i)\}_{i=1}^{n_k}$ represents the $k$-th domain of the $n$-th task. An instance $\mathbf{x}_i$ belongs to class $y_i \in Y_n$, where $Y_n$ is the label space for task $n$, and $Y_n \cap Y_{n'} = \varnothing$ for $n \neq n'$, ensuring that classes are task-specific and non-overlapping, while domains within the same task share the same label spaces. Consequently, during the $n$-th incremental training phase, the model can only access data from $\mathcal{S}_n$. Upon entering the $n$-th incremental inference phase, the model is tasked with joint evaluation on all seen classes $\mathcal{Y}_{n} = \bigcup_{j=1}^n Y_j$. Specifically, the test set at this stage is constructed as $\mathcal{T}_n = \{\mathcal{T}_n^1, \dots, \mathcal{T}_n^{K}, \mathcal{T}_n^{K+1}\}$, where $\mathcal{T}_n^{1:K}$ correspond to seen domains, and $\mathcal{T}_n^{K+1}$ denotes an unseen domain that was entirely excluded from the training trajectory. Our goal is to develop a unified model $h(\mathbf{x}): X \rightarrow \mathcal{Y}_{n}$ that minimizes the empirical risk:
\begin{equation}
h^* = \arg\min_{h \in \mathcal{H}} \mathbb{E}_{(\mathbf{x}, y) \in \mathcal{T}_{1}\cup \cdots \mathcal{T}_{n} } \mathbb{I}(y \neq h(\mathbf{x})),
\end{equation}
where $\mathcal{H}$ denotes the hypothesis space. For implementation, we decompose the model into a non-linear feature extractor $f_{\theta}:\mathcal{X}\rightarrow \mathcal{Z}$ and a linear classifier $g_{\varphi}:\mathcal{Z}\rightarrow\mathbb{R}^{\mathcal{Y}_{n}}$, with parameters $\theta$ and $\varphi$, respectively. The final prediction is given by $h_{\Psi}(\mathbf{x}) = g_{\varphi}(f_{\theta}(\mathbf{x}))$, where $\Psi=(\theta, \varphi)$.

\subsection{Representative Baselines in CIL and DG}
Existing CIL and DG strategies can be summarized into a similar objective:
\begin{equation}
    \min_{\Psi} \sum_{(\mathbf{x},y)\in\mathcal{S}_{n}} \mathcal{L}_{cls}(h_{\Psi} (\mathbf{x}),y)+\mathcal{L}_{reg},
    \label{eq:baseline_loss}
\end{equation}
where $\mathcal{L}_{cls}$ is the standard classification loss and $\mathcal{L}_{reg}$ represents task-specific regularization to address catastrophic forgetting or domain shift.
\paragraph{Baseline in CIL}
In CIL, the objective is to learn representations that are robust across tasks and domains while mitigating catastrophic forgetting. At each incremental step, the model must integrate newly arriving classes and simultaneously retain performance on previously learned ones. In Eq.~\ref{eq:baseline_loss}, $\mathcal{L}_{cls}$ denotes the classification loss for current task samples, while $\mathcal{L}_{reg}$ serves to preserve prior knowledge, \textit{e.g.}, structural constraints~\cite{DBLP:conf/cvpr/YanX021}, rehearsal-based strategies~\cite{aljundi2019gradient}, and knowledge distillation~\cite{DBLP:conf/eccv/DouillardCORV20}. By jointly minimizing both terms, the model is encouraged to encode task-specific information without overwriting past representations as data distributions evolve over time.

\paragraph{Baseline in DG}
In DG, the goal is to extract domain-agnostic and class-discriminative representations that generalize to unseen target domains not available during training. Given multiple labeled source domains, the model learns to capture shared semantics while suppressing domain-specific variations. In Eq.~\ref{eq:baseline_loss}, $\mathcal{L}_{cls}$ is the standard classification loss on seen domains, while $\mathcal{L}_{reg}$ encourages domain invariance, \textit{e.g.}, domain adversarial training~\cite{DBLP:conf/nips/000300M20}, feature alignment~\cite{DBLP:conf/cvpr/LiPWK18}, and contrastive mechanisms~\cite{DBLP:conf/cvpr/Qiao021}. Through this joint objective, the model can generalize beyond seen domains by disentangling semantic content from domain bias.

\paragraph{Discussions}
Although CIL and DG both aim to learn generalizable representations, they are tailored to different and largely disjoint challenges. CIL focuses on mitigating forgetting across evolving class distributions but often lacks mechanisms to handle domain shifts, making it vulnerable to out-of-distribution data. In contrast, DG is designed to address domain shifts but assumes a fixed, complete label space, rendering it ineffective in incremental settings. Consequently, neither approach alone is sufficient for tackling scenarios CDCIL, where both catastrophic forgetting and domain shift occur simultaneously.

\section{Method: Cross-Sample Relational Fusion}
\begin{figure*}[!t]
    \centering
    \includegraphics[width=\linewidth]{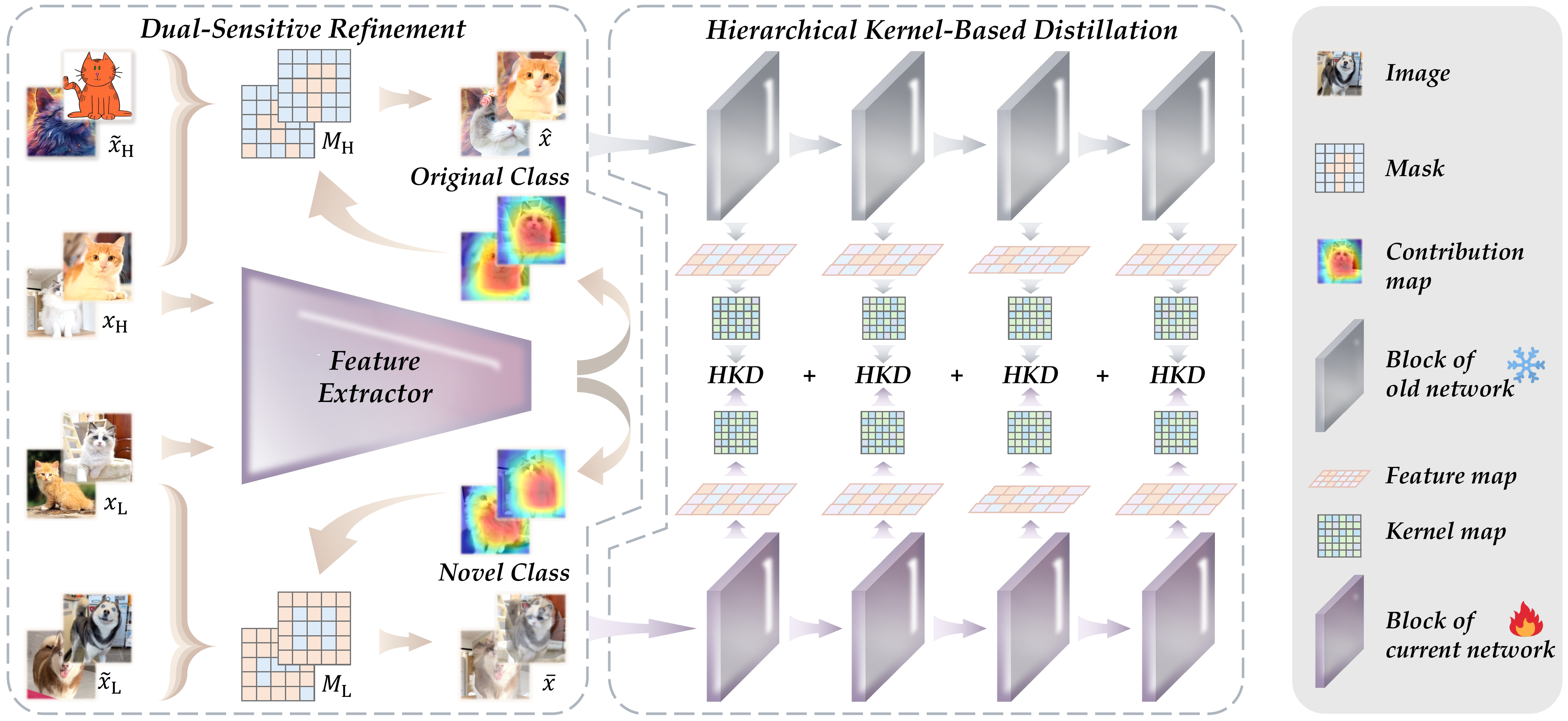}
    \caption{
        Overview of C\ORF.
        \textbf{Left:} Dual-Sensitive Refinement. Given a task-specific dataset $\mathcal{S}_n$, we first select $\mathcal{X}_{\mathrm{high}}$ and $\mathcal{X}_{\mathrm{low}}$ based on confidence scores, and generate corresponding spatial contribution maps. These maps guide the synthesis of an auxiliary dataset $\mathcal{A}_n$ through confidence- and contribution-sensitive blending.
        \textbf{Right:} Hierarchical Kernel-Based Distillation. Given the combined dataset $\mathcal{S}_n \cup \mathcal{A}_n$, feature maps are extracted from both the old and current networks at multiple layers. These maps are then used to construct kernel maps that capture inter-sample structural relations. The HKD between the old and current kernel matrices is computed at each layer and aggregated to form the final distillation loss.
            }
    \label{fig:overview}
\end{figure*}

We propose C\ORF\ to address the combined challenges of forgetting and domain shift in CDCIL by learning domain-agnostic, incrementally stable representations. As shown in Fig.~\ref{fig:overview}, C\ORF\ consists of two synergistic components. Dual-Sensitive Refinement (DSR) synthesizes an auxiliary dataset and jointly trains it with the original training set, encouraging the model to form compact intra-class clusters and separable inter-class boundaries in a domain-invariant space. To further strengthen learned representations and preserve structural consistency, we introduce Hierarchical Kernel-Based Distillation (HKD), which captures fine-grained and domain-sensitive structure. These components complement each other by enhancing the quality of representations and preserving structural knowledge across tasks and domains, enabling effective continual learning under domain shift.

\subsection{Dual-Sensitive Refinement (DSR)} \label{sub:dsr}
To build a domain-agnostic representation space in CDCIL, the model is required to align semantically consistent instances across heterogeneous seen domains while preserving inter-class separability under unseen domains. However, conventional cross-entropy objectives, designed for stationary single-domain settings, often fail to generalize, as they neglect domain-induced biases and overfit to spurious contextual artifacts. To address this, we strategically leverage spatial contribution and predictive confidence to jointly construct a domain-agnostic representation space and stabilize features, thereby mitigating catastrophic forgetting.

\paragraph{High-Confidence Blending}
For each batch of samples $\{\mathbf{x}_i\}_{i=1}^B$, we identify high-confidence samples by selecting those with the highest predicted probabilities. These samples are assumed to contain more reliable semantic structures and less noise, thus serving as robust anchors for representation learning. Specifically, given a batch $\{\mathbf{x}_i\}_{i=1}^B$, we compute the predictive confidence for each sample $\mathbf{x}_i$ as ${p_i = \max_j g_j(f(\mathbf{x}_i))}$, where $g_j(\cdot)$ denotes the normalized output for class $j$. We then select the top-$k$ samples with the highest predictive confidence $p_i$:
\begin{equation} \label{eq:3}
    \mathcal{X}_{\mathrm{high}} = \left\{ \mathbf{x}_i \mid p_i \in \mathrm{Top}_k(p_i) \right\}.
\end{equation}
For each high-confidence sample $\mathbf{x}_{\mathrm{H}} \in \mathcal{X}_{\mathrm{high}}$, we retrieve a same-class counterpart $\tilde{\mathbf{x}}_{\mathrm{H}}$ from a different seen domain to introduce cross-domain diversity while preserving the class label. 
To decouple semantic cues from domain-specific patterns, we employ Grad-CAM~\cite{DBLP:conf/iccv/SelvarajuCDVPB17} on the last convolutional block of the backbone (\textit{i.e.}, the final feature map before global average pooling) to compute a spatial contribution map for $\mathbf{x}_{\mathrm{H}}$. Note that Grad-CAM is enabled from the beginning but is only used for region selection rather than direct supervision, and its influence is naturally bounded by the confidence-based top-$k$ sampling. We then binarize this map by assigning 1 to the top $\alpha$ proportion of values and 0 to the rest, resulting in a binary mask \(\mathbf{M}_{\mathrm{H,H}}\) that highlights the most class-discriminative regions. These regions correspond to the areas that contribute most significantly to the model’s prediction. The complementary mask, defined as \(\mathbf{M}_{\mathrm{H,L}} = \mathbf{1} - \mathbf{M}_{\mathrm{H,H}}\), captures regions with low attribution, which are more likely to contain domain-specific noise or background information that is less relevant to the classification task.

Using these masks, we blend the low-contribution regions of $\mathbf{x}_{\mathrm{H}}$ and $\tilde{\mathbf{x}}_{\mathrm{H}}$ while preserving the highly activated regions of $\mathbf{x}_{\mathrm{H}}$ to synthesize the fused sample:
\begin{equation} \label{eq:4}
\hat{\mathbf{x}} = \mathbf{M}_\mathrm{H, H} \odot \mathbf{x}_\mathrm{H} + \frac{\mathbf{M}_\mathrm{H, L}}{2} \odot \left(\mathbf{x}_\mathrm{H} + \tilde{\mathbf{x}}_\mathrm{H} \right),
\end{equation}
where $\odot$ denotes element-wise multiplication. These synthesized samples are predicted as the same class as $\mathbf{x}_{\mathrm{H}}$. This design encourages the model to learn from semantically reliable, high-confidence regions while remaining robust to noise in diverse, low-confidence regions blended across domains, thereby improving class-consistent representations and reducing dependence on domain-specific cues.

\paragraph{Low-Confidence Blending}
In contrast to that of high-confidence instances, low-confidence samples are identified as those with the lowest predicted probabilities within a mini-batch. These samples typically contain ambiguous semantic patterns, domain-induced distortions, or weak class-discriminative cues. Formally, given a batch $\{\mathbf{x}_i\}_{i=1}^B$, we define the low-confidence set $\mathcal{X}_{\mathrm{low}}$ :
\begin{equation} \label{eq:5}
\mathcal{X}_{\mathrm{low}} = \left\{ \mathbf{x}_i \mid p_i \in \mathrm{Bottom}_k(p_i) \right\}.
\end{equation}
For each low-confidence sample $\mathbf{x}_{\mathrm{L}} \in \mathcal{X}_{\mathrm{low}}$, we pair it with a counterpart $\tilde{\mathbf{x}}_{\mathrm{L}}$ from a different class within the same batch to encourage feature disentanglement.

We then compute the contribution maps $\mathbf{M}_{\mathrm{L,H}}$ and $\mathbf{M}_{\mathrm{L,L}}$ for $\mathbf{x}_{\mathrm{L}}$, which quantify the spatial contribution of each region to the model’s prediction. Guided by these maps, $\mathbf{x}_{\mathrm{L}}$ is fused with $\tilde{\mathbf{x}}_{\mathrm{L}}$ as follows:
\begin{equation} \label{eq:6}
    \bar{\mathbf{x}}= \frac{\mathbf{M}_\mathrm{L, H}}{2} \odot \left(\mathbf{x}_\mathrm{L} + \tilde{\mathbf{x}}_\mathrm{L} \right) + \mathbf{M}_\mathrm{L, L} \odot \mathbf{x}_\mathrm{L}.
\end{equation}
These synthesized samples are assigned to auxiliary classes that are disjoint from the original label space~\cite{DBLP:conf/nips/ZhuCZL21}. Conceptually, auxiliary categories correspond to pairwise class combinations. 
However, in practice we do not enumerate all possible class pairs. 
Auxiliary labels are instantiated only for the blended low-confidence samples within each mini-batch, and the number of such samples is bounded by the bottom-$k$ selection budget ($k_{\mathrm{low}}=\alpha B$). 
We reuse a fixed auxiliary label index set across iterations, so auxiliary labels serve as temporary supervision placeholders rather than permanent semantic classes. 
This design prevents semantic ambiguity and avoids corrupting the original classification boundaries without introducing quadratic growth in the label space.
In early stages, potential noise is mitigated since the blending is only applied to a small bottom-$k$ subset and the resulting samples are assigned to auxiliary classes, avoiding unstable gradients on the original label space.

\paragraph{{Effect of Dual-Sensitive Refinement}}
The synthesized samples $\hat{\mathbf{x}}$ and $\bar{\mathbf{x}}$ generated by DSR are organized to construct an auxiliary set $\mathcal{A}_n$, which serves as a robust anchor for constructing a domain-agnostic representation space. 
In practice, the auxiliary set is dynamically formed within each mini-batch using only the selected high- and low-confidence samples, and its size is bounded by the selection budget $k_{\mathrm{low}}=\alpha B$. 
We reuse a fixed auxiliary label space throughout training, thereby preventing combinatorial growth of auxiliary categories across incremental stages. Auxiliary-labeled blended samples are generated on-the-fly and are not stored in memory; when integrated with replay-based baselines, exemplar selection and replay follow the original protocol and involve only real samples with their original class labels.
To promote generalization across domains, we optimize a unified objective jointly over original and synthesized samples:
\begin{equation} 
    \mathrm{\min} \sum_{(\mathbf{x},y)\in\mathcal{S}_{n} \cup \mathcal{A}_{n}} \mathcal{L}_{cls}(h_{\Psi} (\mathbf{x}),y)+\mathcal{L}_{reg}.
    \label{eq:dsr_loss}
\end{equation}
As illustrated in Fig.~\ref{fig:overview} (Left), DSR adaptively calibrates learning signals at both spatial and instance levels through contribution- and confidence-aware blending. This selective mixing process effectively suppresses domain-specific artifacts while preserving task-relevant structures, thereby facilitating the construction of a domain-agnostic representation space. By guiding the model to attend to transferable and discriminative patterns, DSR enhances robustness and generalizability across incrementally encountered domains.

\paragraph{DSR vs. CutMix}
Although DSR shares superficial similarity with CutMix-style mixing~\cite{zhang2017mixup,yun2019cutmix}, it is contribution-guided rather than random: Grad-CAM masks localize class-relevant regions and restrict fusion mainly to low-contribution areas that often encode domain bias in CDCIL.
In addition, DSR handles low-confidence samples via auxiliary-class assignment to prevent boundary corruption, which goes beyond label-proportional mixing in CutMix.

\subsection{Hierarchical Kernel-Based Distillation (HKD)}\label{sub:hkd}
While DSR facilitates the construction of a domain-agnostic representation space, it should be complemented by a robust strategy for preserving knowledge over time. Existing methods predominantly employ logit-level knowledge distillation~\cite{DBLP:journals/corr/HintonVD15}. However, such approaches typically assume distributional consistency between stages, an assumption that rarely holds under domain shift. Consequently, the supervision signals become biased toward dominant domains, leading to semantic drift and weakening long-term knowledge retention.

\paragraph{Relational Distillation via Kernel Similarity}
To mitigate catastrophic forgetting, we propose hierarchical kernel-based distillation (HKD), which reframes knowledge distillation as a fine-grained, geometry-aware relational alignment task. Rather than relying solely on high-level logits or global representations, HKD captures the \textit{sample relations} at multiple network depths to preserve structural knowledge, thereby effectively alleviating catastrophic forgetting. The old model denotes the network from the previous stage that preserves prior knowledge, whereas the current model refers to the network under training for acquiring new tasks without forgetting.

Specifically, let $(l_t, l_s) \in \mathcal{R}$ denote a pair of corresponding layers from the old and current model. For an input sample $\mathbf{x}_i$, we extract the flattened feature maps $q^{l_t}(\mathbf{x}_i)$ and $q^{l_s}(\mathbf{x}_i)$. Then we construct the \textit{relational kernel map} $\mathcal{P}^{l_t} = \{ p_{(i,j)}^{l_t} \}_{i,j=1}^{B}$ for the old model and analogously $\mathcal{P}^{l_s}$ for the current model. These maps characterize the relational topology within the batch, capturing the underlying geometric relationships between samples beyond their individual activations:
\begin{align} \label{eq:8}
    \begin{aligned}
        p_{(i,j)}^{l_t}=\frac{\mathcal{K}(q^{l_t}(\mathbf{x}_i), q^{l_t}(\mathbf{x}_j))}{\sum_{i=1, i\neq j}^B \mathcal{K}(q^{l_t}(\mathbf{x}_i), q^{l_t}(\mathbf{x}_j))},\\
        p_{(i,j)}^{l_s}=\frac{\mathcal{K}(q^{l_s}(\mathbf{x}_i), q^{l_s}(\mathbf{x}_j))}{\sum_{i=1, i\neq j}^B \mathcal{K}(q^{l_s}(\mathbf{x}_i), q^{l_s}(\mathbf{x}_j))},
    \end{aligned}
\end{align}
where $\mathcal{K}(\cdot, \cdot)$ denotes a similarity kernel, and $B$ is the batch size. Here, $p_{(i,j)}^{l_t}$ and $p_{(i,j)}^{l_s}$ represent the conditional similarity between $\mathbf{x}_i$ and $\mathbf{x}_j$ within the old representation spaces and the current representation spaces, respectively.

The choice of kernel function critically affects the fidelity of the relational structure. We adopt a hybrid kernel design to leverage complementary geometric properties, namely cosine  kernel $\mathcal{K}_C(\mathbf{a}, \mathbf{b}) = \frac{1}{2} \left( \frac{\mathbf{a}^\top \mathbf{b}}{\|\mathbf{a}\|_2 \|\mathbf{b}\|_2} + 1 \right)$ and student-$t$ kernel $\mathcal{K}_T(\mathbf{a}, \mathbf{b}) = \frac{1}{1 + \|\mathbf{a} - \mathbf{b}\|_2}$. The former emphasizes angular alignment, critical for preserving semantic directions in high-dimensional spaces, while the latter captures localized structures and is robust to outliers.

\paragraph{Symmetric Divergence-Based Alignment}
To align the relational distributions between the old and current model, we define the alignment target $\mathcal{D}(\mathcal{P}^{l_t} | \mathcal{P}^{l_s})$ as:
\begin{align} ~\label{eq:target}
    - \sum_{i=1}^B \sum_{\substack{j=1, j \neq i}}^B
    \left(p_{(i,j)}^{l_t} - p_{(i,j)}^{l_s}\right) \cdot \left(\log p_{(i,j)}^{l_t} - \log p_{(i,j)}^{l_s}\right){.}
\end{align}

For each kernel variant, we compute the divergence independently, yielding the distillation objective at layer pair $(l_t, l_s)$:
\begin{equation} \label{eq:9}
\mathcal{L}^{l_t, l_s} = \mathcal{D}(\mathcal{P}_C^{l_t} \| \mathcal{P}_C^{l_s}) + \mathcal{D}(\mathcal{P}_T^{l_t} \| \mathcal{P}_T^{l_s}),
\end{equation}
where $C$ and $T$ denote the similarity matrices computed using the kernels $\mathcal{K}_C(\cdot,\cdot)$ and $\mathcal{K}_T(\cdot,\cdot)$, respectively.
\paragraph{Hierarchical Aggregation}
We aggregate the distillation losses over selected layer pairs to obtain the overall hierarchical distillation objective:
\begin{equation} \label{eq:10}
\mathcal{L}_{kernel} = \sum_{(l_t, l_s) \in \mathcal{R}} \mathcal{L}^{l_t, l_s}. 
\end{equation}
This enforces topological consistency across the representation hierarchy, enabling the current model to faithfully preserve fine-grained, domain-sensitive relational structures while evolving through continual adaptation.

\paragraph{Effect of Hierarchical Kernel-Based Distillation}
As illustrated in Fig.~\ref{fig:overview} (Right), by aligning kernel-induced relational topologies across network depths, HKD enables the current network to internalize structural knowledge without relying on explicit labels or task boundaries. It effectively addresses catastrophic forgetting by preserving class-discriminative and domain-invariant sample relations, offering a geometry-grounded and scalable solution for knowledge retention in CDCIL.

\begin{algorithm}[ht]
    \caption{Training C\ORF\ for CDCIL}
    \label{alg:1}
    \raggedright
    {\bf Input}: 
    Training dataset: $\mathcal{S}_n = \{\mathcal{D}^1_n, \mathcal{D}^2_n, \dots, \mathcal{D}^{K}_n\}$;
    old model $f_{\text{old}}(\cdot)$; 
    current model $f_{\text{new}}(\cdot)$ ;\\
    {\bf Output}: Updated model;
    \begin{algorithmic}[1]
        \State Get current training set $\mathcal{S}_n = \{\mathcal{D}^1_n, \mathcal{D}^2_n, \dots, \mathcal{D}^{K}_n\}$;
        
\noindent\rule{\linewidth}{0.4pt} \textbf{DSR}:
        \For{batch $\{(\mathbf{x}_i, y_i)\}_{i=1}^B \in \mathcal{S}_n$}
            \State Calculate predictive confidence $\{p_i\}_{i=1}^B$; \label{line:3}
            \State Identify confidence sets $\mathcal{X}_{\mathrm{high}}$,$\mathcal{X}_{\mathrm{low}}$ via Eq.~\ref{eq:3},~\ref{eq:5}; \label{line:4}
            \For{$\mathbf{x}_{\mathrm{H}} \in \mathcal{X}_{\mathrm{high}}$, $\mathbf{x}_{\mathrm{L}} \in \mathcal{X}_{\mathrm{low}}$}
            {\Comment{\color{cyan}{Synthesize sample}}}
                \State Select $\tilde{\mathbf{x}}_{\mathrm{H}}$ (same class, different domain) and $\tilde{\mathbf{x}}_{\mathrm{L}}$ (different class, same batch); \label{line:6}
                \State Calculate spatial contribution map for $\mathbf{x}_{\mathrm{H}}$, $\mathbf{x}_{\mathrm{L}}$;
                \State Binarize contribution map to obtain $\mathbf{M}_{\mathrm{H,H}}$, $\mathbf{M}_{\mathrm{L,L}}$;
                \State Calculate complementary masks $\mathbf{M}_{\mathrm{H,L}}=\mathbf{1}-\mathbf{M}_{\mathrm{H,H}}$, $\mathbf{M}_{\mathrm{L,H}}=\mathbf{1}-\mathbf{M}_{\mathrm{L,L}}$;\\
                {\Comment{\color{cyan}{Complementary mask}}}
                \State Synthesize fused sample $\hat{\mathbf{x}}$, $\bar{\mathbf{x}}$ via Eq.~\ref{eq:4},~\ref{eq:6}; \label{line:10}
            \EndFor
        \EndFor
     
        \State Construct auxiliary dataset $\mathcal{A}_n$ from $\{\hat{\mathbf{x}}\}$ and $\{\bar{\mathbf{x}}\}$;
        \State Calculate the loss via Eq.~\ref{eq:dsr_loss};\label{line:14}

\noindent\rule{\linewidth}{0.4pt} \noindent\textbf{HKD}:
        \For{layer pair $(l_{\text{t}} \in f_{\text{old}}(\cdot),\; l_{\text{s}} \in f_{\text{new}}(\cdot))$}
            \For{$(\mathbf{x}, y) \in \mathcal{S}_n \cup \mathcal{A}_n$}
            {\Comment{\color{cyan}{Incremental learning}}}
                \State Extract feature maps $q^{l_t}(\mathbf{x})$, $q^{l_s}(\mathbf{x})$; \label{line:17}
                \State Construct kernel maps $\mathcal{P}^{l_t}$, $\mathcal{P}^{l_s}$ via Eq.~\ref{eq:8} using $\mathcal{K}_C$ and $\mathcal{K}_T$; \label{line:18}
                \State Calculate $\mathcal{D}(\mathcal{P}^{l_t} \| \mathcal{P}^{l_s})$; 
                {\Comment{\color{cyan}{Symmetric alignment}}} \label{line:19}
                \State Align relational distributions $\mathcal{L}^{l_t, l_s}$ via Eq.~\ref{eq:9}; \label{line:20}
            \EndFor
        \EndFor
        \State Aggregate distillation losses $\mathcal{L}_{kernel}$ via Eq.~\ref{eq:10}; \label{line:23}
\noindent\rule{\linewidth}{0.4pt}\State Update the model via Eq.~\ref{eq:final_loss}; \label{line:24}
        
 \Return{The updated model};
    \end{algorithmic}
\end{algorithm}

\subsection{Summary for C\ORF}
We formulate C\ORF\ as a unified optimization objective:
\begin{equation} 
    \mathrm{\min} \sum_{(\mathbf{x},y)\in\mathcal{S}_{n} \cup \mathcal{A}_{n}} \mathcal{L}_{cls}(h_{\Psi} (\mathbf{x}),y)+\mathcal{L}_{reg}+\beta \cdot \mathcal{L}_{kernel},
    \label{eq:final_loss}
\end{equation}
where $\beta$ is the balancing coefficient. Compared to Eq.~\ref{eq:baseline_loss}, we extend the baseline in two complementary directions: \textbf{1)} at the data level, DSR introduces the auxiliary set $\mathcal{A}_n$, which is directly merged with the original training set $\mathcal{S}_n$ to provide additional domain-diverse samples during optimization; and \textbf{2)} at the structural level, HKD contributes the extra supervision $\mathcal{L}_{kernel}$, which is seamlessly incorporated alongside the conventional regularization $\mathcal{L}_{reg}$ to enforce multi-grained alignment for mitigating catastrophic forgetting. As illustrated in Fig.~\ref{fig:overview}, these two components in C\ORF\ work synergistically to form a unified learning paradigm: data-level DSR fosters domain-invariant and generalizable representations, while structure-level HKD preserves relational continuity, together enabling the model to build a domain-agnostic and incrementally robust representation space for CDCIL.

\noindent\textbf{Pseudo Code:} We provide the pseudo-code of C\ORF\ to illustrate the training process in Alg.~\ref{alg:1}. First, we are provided with the training dataset $S_n$ and the old model $f_{\text{old}}(\cdot)$ to update the current model $f_{\text{new}}(\cdot)$. 
The process begins with the \textbf{DSR} phase. Specifically, predictive confidence is calculated for all samples in the dataset (Line~\ref{line:3}). Based on these predictive confidence, the samples are identified into the high-confidence and low-confidence samples subsets
(Line~\ref{line:4}). For each subset, these samples are used to calculate spatial contribution maps and to synthesize fused samples (Lines~\ref {line:6} to~\ref {line:10}). These synthesized samples are then used to construct the auxiliary dataset $\mathcal{A}_n = \{\hat{\mathbf{x}}, \bar{\mathbf{x}}\}$.

Next, we enter the \textbf{HKD} phase. For each layer pair ($l_{\text{t}}$, $l_{\text{s}}$) from the old and new models, the corresponding feature maps are extracted and transformed into kernel representations using the cosine
similarity kernel $\mathcal{K}_C(\cdot, \cdot)$ and the student-$t$ kernel $\mathcal{K}_T(\cdot, \cdot)$ (Line~\ref{line:17} to Line~\ref{line:18}). These kernel maps are then used to construct relational distributions, upon which an alignment target is calculated (Line~\ref{line:19} to Line~\ref{line:20}). Finally, the distillation losses are aggregated (Line~\ref{line:23}), and the model is updated based on the loss function  (Line~\ref{line:24}). The updated model is then
returned as the output of the training process.

\noindent\textbf{Applicability to Backbone Architectures:}
C\ORF\ is designed for CIL frameworks where backbone parameters are updated during incremental training. 
Many recent ViT-based CIL methods freeze the pretrained backbone and update only lightweight modules such as prompts or adapters. 
Since both DSR and HKD operate by reshaping and aligning intermediate feature representations during backbone optimization, 
their effectiveness relies on trainable backbone parameters. 
Therefore, the current formulation of C\ORF\ is primarily suited for convolutional backbones.

\section{Experiments}
\label{sec:experiments}
We evaluate C\ORF\ on three benchmark datasets to demonstrate its seamless integration with other CIL baselines, followed by ablation study, sensitivity analysis, and other experiments to verify its effectiveness and robustness.

\begin{figure*}[t]
    \centering
    \includegraphics[width=\linewidth]{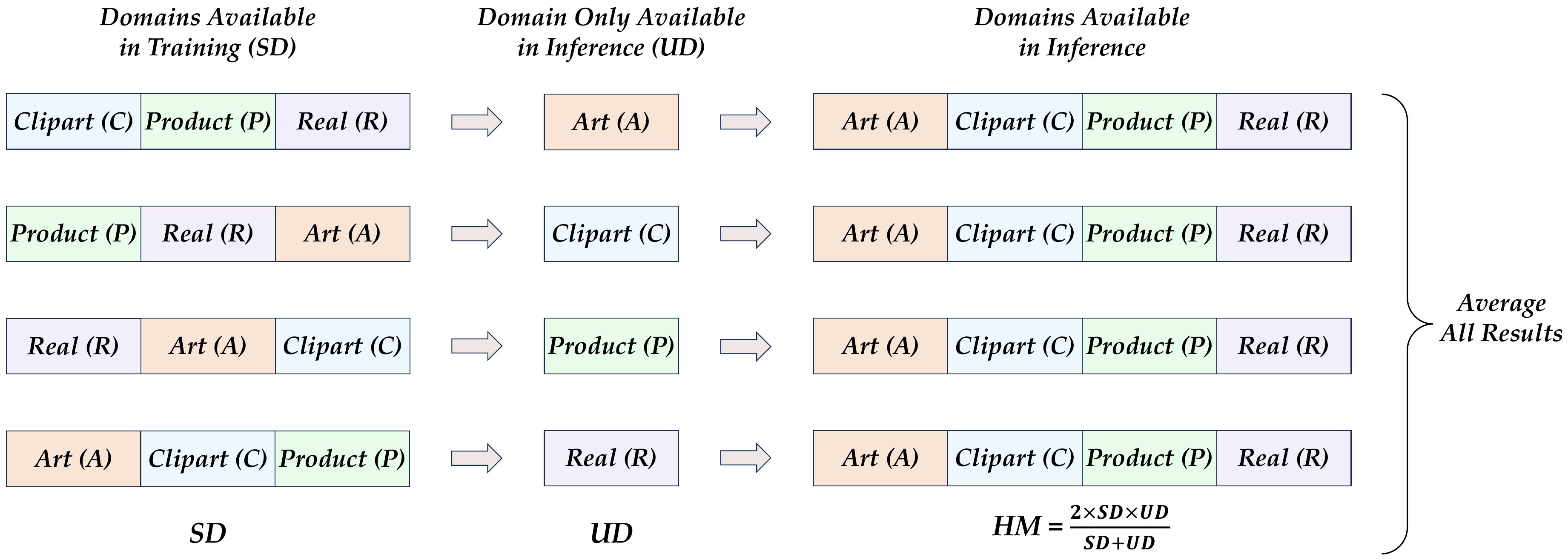}
    \caption{Evaluation protocol for CDCIL on OfficeHome. 
Each domain is treated once as the unseen domain (UD), while the others serve as seen domains (SD). 
Final-stage SD, UD, and HM are computed and averaged over all domain rotations.}
    \label{fig:EvaluationProtocol}
\end{figure*}

\subsection{Implementation Details}
\paragraph{Datasets}
Following \cite{DBLP:conf/cvpr/SimonFT0SSHC22} and \cite{DBLP:journals/air/KhoeeYF24}, we evaluate the proposed framework on three datasets: OfficeHome~\cite{DBLP:conf/cvpr/VenkateswaraECP17}, DomainNet~\cite{DBLP:conf/iccv/SaitoKSDS19}, and PACS~\cite{DBLP:conf/iccv/LiYSH17}. These datasets are selected for their diversity in visual styles and domain shifts. We adopt a class-split strategy that ensures the number of classes is as balanced as possible across tasks. For OfficeHome, the 65 classes are equally partitioned into 13 tasks and 5 tasks. For DomainNet, the 345 classes are equally divided into 5 tasks. For PACS, training begins with 3 base classes and proceeds with incremental tasks of 2 classes each. For methods that rely on exemplars, we allocate a fixed exemplar budget of 20 for OfficeHome and PACS, and 10 for DomainNet. To ensure fairness, consistency, and reproducibility, we fix the random seed to 1993 for the entire experimental pipeline, including data shuffling, task splitting and model initialization~\cite{DBLP:conf/cvpr/RebuffiKSL17}.

\paragraph{Training Details}
All experiments are implemented using PyTorch~\cite{DBLP:conf/nips/PaszkeGMLBCKLGA19} and trained on an NVIDIA RTX 4090 GPU. We employ ResNet-34~\cite{DBLP:conf/cvpr/HeZRS16} as the backbone network, trained from scratch across all tasks for a fair comparison. In C\ORF, we use a batch size of $200$ and train for $200$ epochs using the SGD optimizer with momentum fixed at $0.9$. The initial learning rate is set to $0.1$ and decayed using a cosine annealing schedule. Weight decay is set to $5e-4$. In DSR, the same hyperparameter $\alpha$ is used for both Top-$k$ selection and binarizing the contribution map, where $k = \alpha \cdot B$. We fix $\alpha = 0.3$ and $\beta = 0.01$ throughout all experiments.

\begin{table*}[!t]
	\caption{
    Average accuracy on seen domains (SD), unseen domain (UD), and their harmonic mean (HM) across three datasets using ResNet-34 as the backbone. The best results are highlighted in bold.
    }
	\label{tab:benchmark}
	\centering
	\resizebox{1.0\textwidth}{!}{%
		\begin{tabular}{@{}l cccc cccc cccc}
			\toprule
			\multicolumn{1}{l}{\multirow{2}{*}{Method}} & 
			\multicolumn{3}{c}{Officehome 13 tasks} & \multicolumn{3}{c}{Officehome 5 tasks}
			& \multicolumn{3}{c}{DomainNet 5 tasks}
			& \multicolumn{3}{c}{PACS 3 tasks}
			 \\
			& {$\bar{A}-\mathrm{SD}$} & {$\bar{A}-\mathrm{UD}$}  
			& {$\bar{A}-\mathrm{HM}$}
			& {$\bar{A}-\mathrm{SD}$} & {$\bar{A}-\mathrm{UD}$}  
			& {$\bar{A}-\mathrm{HM}$}
			& {$\bar{A}-\mathrm{SD}$} & {$\bar{A}-\mathrm{UD}$}  
			& {$\bar{A}-\mathrm{HM}$}
			& {$\bar{A}-\mathrm{SD}$} & {$\bar{A}-\mathrm{UD}$}  
			& {$\bar{A}-\mathrm{HM}$}
			\\
			\midrule
			\rowcolor{gray!10}  MSL+MOV & 29.13 ± {\footnotesize 4.28} & 16.22 ± {\footnotesize 5.43} & 18.94 ± {\footnotesize 3.85}  & 31.13 ± {\footnotesize 2.98} & 15.25 ± {\footnotesize 5.83} & 19.63 ± {\footnotesize 6.20} & 21.04 ± {\footnotesize 1.53} & 16.33 ± {\footnotesize 6.92} & 17.02 ± {\footnotesize 6.38} &  45.89 ± {\footnotesize 1.73} & 20.82 ± {\footnotesize 9.57} & 26.97 ± {\footnotesize 9.84} \\
			\midrule
			\rowcolor{gray!10}  TRIPS & 31.56 ± {\footnotesize 4.98} & 17.84 ± {\footnotesize 4.32} & 20.71 ± {\footnotesize 5.76}  & 36.82 ± {\footnotesize 5.12} & 17.97 ± {\footnotesize 5.47} & 23.77 ± {\footnotesize 4.49} & 24.38 ± {\footnotesize 2.18} & 15.26 ± {\footnotesize 6.39} & 16.94 ± {\footnotesize 5.92} &  49.47 ± {\footnotesize 3.87} & 21.51 ± {\footnotesize 16.38} & 28.50 ± {\footnotesize 11.62} \\
			\midrule
			FineTune& 17.20 ± {\scriptsize 0.76} & 13.31 ± {\scriptsize 1.91} & 14.77 ± {\scriptsize 1.19} & 29.24 ± {\scriptsize 2.06} & 19.00 ± {\scriptsize 3.49} & 22.65 ± {\scriptsize 2.16} & 34.53 ± {\scriptsize 1.31} & 22.90 ± {\scriptsize 8.30} & 26.47 ± {\scriptsize 6.38} & 52.26 ± {\scriptsize 1.43} & 37.08 ± {\scriptsize 9.78} & 42.04 ± {\scriptsize 7.84} \\
			\midrule
			\rowcolor{LightCyan}FineTune w/ C\ORF & \textbf{19.15 ± {\scriptsize 0.84}} & \textbf{15.11 ± {\scriptsize 0.59}} & \textbf{16.49 ± {\scriptsize 0.07}} & \textbf{30.90 ± {\scriptsize 2.37}} & \textbf{23.33 ± {\scriptsize 1.10}} & \textbf{25.86 ± {\scriptsize 0.06}} & \textbf{34.99 ± {\scriptsize 0.24}} & \textbf{26.26 ± {\scriptsize 8.90}} & \textbf{28.96 ± {\scriptsize 6.90}} & \textbf{52.95 ± {\scriptsize 1.44}} & \textbf{43.90 ± {\scriptsize 2.68}} & \textbf{47.46 ± {\scriptsize 1.11}} \\
			\midrule
			Replay & 
            48.30 ± {\scriptsize 3.78} & 
            29.72 ± {\scriptsize 6.76} & 
            35.74 ± {\scriptsize 4.14}  &  
            48.25 ± {\scriptsize 4.00} & 
            28.63 ± {\scriptsize 6.02} & 
            35.07 ± {\scriptsize 3.78} & 
            41.79 ± {\scriptsize 1.75} & 
            26.12 ± {\scriptsize 9.97} & 
            30.72 ± {\scriptsize 7.97} &  
            64.21 ± {\scriptsize 3.93} & 
            40.44 ± {\scriptsize 10.93} & 
            48.15 ± {\scriptsize 10.20} \\
			\midrule
			\rowcolor{LightCyan}Replay w/ C\ORF & 
            \textbf{50.10 ± {\scriptsize 4.30}} & 
            \textbf{31.65 ± {\scriptsize 7.07}} & 
            \textbf{37.70 ± {\scriptsize 4.29}} & 
            \textbf{51.51 ± {\scriptsize 4.35}} & 
            \textbf{31.23 ± {\scriptsize 6.43}} & 
            \textbf{38.02 ± {\scriptsize 3.91}} & 
            \textbf{44.56 ± {\scriptsize 1.70}} & 
            \textbf{30.98 ± {\scriptsize 8.90}} & 
            \textbf{35.48 ± {\scriptsize 6.55}} & 
            \textbf{68.90 ± {\scriptsize 4.40}} & 
            \textbf{49.70 ± {\scriptsize 11.20}} & 
            \textbf{56.74 ± {\scriptsize 8.31}} \\
			\midrule
			iCaRL & 49.41 ± {\scriptsize 3.53} & 29.55 ± {\scriptsize 6.52} & 35.95 ± {\scriptsize 4.48} & 49.81 ± {\scriptsize 4.21} & 29.20 ± {\scriptsize 6.57} & 35.90 ± {\scriptsize 4.44} & 45.98 ± {\scriptsize 2.08} & 28.06 ± {\scriptsize 11.03} & 33.22 ± {\scriptsize 8.90} & 68.14 ± {\scriptsize 4.18} & 44.80 ± {\scriptsize 10.19} & 52.59 ± {\scriptsize 9.06} \\
			\midrule
			\rowcolor{LightCyan}iCaRL w/ C\ORF & \textbf{50.77 ± {\scriptsize 4.68}} & \textbf{32.89 ± {\scriptsize 6.76}} & \textbf{38.95 ± {\scriptsize 4.02}}  & \textbf{54.39 ± {\scriptsize 4.71}} & \textbf{33.58 ± {\scriptsize 7.60}} & \textbf{40.55 ± {\scriptsize 4.76}} & \textbf{47.83 ± {\scriptsize 1.95}} & \textbf{32.39 ± {\scriptsize 10.81}} & \textbf{36.99 ± {\scriptsize 8.85}} & \textbf{71.77 ± {\scriptsize 4.35}} & \textbf{50.54 ± {\scriptsize 12.31}} & \textbf{57.89 ± {\scriptsize 9.66}} \\
			\midrule
			DER & 57.39 ± {\scriptsize 4.31} & 
            35.09 ± {\scriptsize 8.04} & 
            42.41 ± {\scriptsize 5.75} & 
            55.09 ± {\scriptsize 5.11} & 
            32.11 ± {\scriptsize 7.57} & 
            39.53 ± {\scriptsize 5.02} & 
            50.66 ± {\scriptsize 2.28} & 
            30.97 ± {\scriptsize 12.96} & 
            36.37 ± {\scriptsize 10.59} & 
            68.65 ± {\scriptsize 3.08} & 
            46.11 ± {\scriptsize 11.57} & 
            53.74 ± {\scriptsize 9.66} \\
			\midrule
			\rowcolor{LightCyan}DER w/ C\ORF & \textbf{60.01 ± {\scriptsize 6.32}} & \textbf{38.54 ± {\scriptsize 6.19}} & \textbf{46.17 ± {\scriptsize 4.25}}  & \textbf{57.92 ± {\scriptsize 5.14}} & \textbf{33.82 ± {\scriptsize 8.06}} & \textbf{41.65 ± {\scriptsize 5.50}} & \textbf{53.36 ± {\scriptsize 0.60}} & \textbf{42.95 ± {\scriptsize 6.64}} & \textbf{47.24 ± {\scriptsize 3.90}} & \textbf{71.37 ± {\scriptsize 4.92}} & \textbf{49.98 ± {\scriptsize 8.62}} & \textbf{57.77 ± {\scriptsize 6.59}}\\
			\midrule
			MEMO &   53.33 ± {\scriptsize 5.22} & 32.73 ± {\scriptsize 7.41} & 39.45 ± {\scriptsize 4.97}   & 53.02 ± {\scriptsize 5.42} & 30.76 ± {\scriptsize 6.87} & 37.97 ± {\scriptsize 5.00} & 46.73 ± {\scriptsize 2.04} & 28.44 ± {\scriptsize 11.65} & 33.56 ± {\scriptsize 9.57}  &  64.65 ± {\scriptsize 1.16} & 42.61 ± {\scriptsize 11.29} & 49.64 ± {\scriptsize 9.93} \\
			\midrule
			\rowcolor{LightCyan}MEMO w/ C\ORF & \textbf{55.15 ± {\scriptsize 5.57}} & \textbf{33.19 ± {\scriptsize 6.75}} & \textbf{40.30 ± {\scriptsize 4.23}}   &   \textbf{56.02 ± {\scriptsize 5.83}} & \textbf{33.25 ± {\scriptsize 6.60}} & \textbf{40.79 ± {\scriptsize 3.74}} & \textbf{51.17 ± {\scriptsize 1.28}} & \textbf{43.47 ± {\scriptsize 7.25}} & \textbf{46.70 ± {\scriptsize 5.01}} &   \textbf{69.99 ± {\scriptsize 1.94}} & \textbf{48.64 ± {\scriptsize 12.26}} & \textbf{55.87 ± {\scriptsize 9.51}}\\
			\midrule
			FOSTER &   53.05 ± {\scriptsize 3.97} & 32.36 ± {\scriptsize 8.19} & 38.98 ± {\scriptsize 5.71}  &   56.08 ± {\scriptsize 4.84} & 33.15 ± {\scriptsize 6.78} & 40.68 ± {\scriptsize 3.76} & 58.30 ± {\scriptsize 2.44} & 36.30 ± {\scriptsize 14.66} & 42.54 ± {\scriptsize 11.96} & 72.30 ± {\scriptsize 4.22} & 45.49 ± {\scriptsize 15.82} & 53.12 ± {\scriptsize 15.38} \\
			\midrule
			\rowcolor{LightCyan}FOSTER w/ C\ORF &   \textbf{53.09 ± {\scriptsize 2.66}} & \textbf{37.94 ± {\scriptsize 2.81}} & \textbf{43.66 ± {\scriptsize 1.81}}  &  \textbf{56.71 ± {\scriptsize 5.38}} & \textbf{35.28 ± {\scriptsize 6.92}} & \textbf{42.49 ± {\scriptsize 3.89}} & \textbf{60.40 ± {\scriptsize 3.54}} & \textbf{40.60 ± {\scriptsize 12.12}} & \textbf{47.16 ± {\scriptsize 9.28}} & \textbf{76.21 ± {\scriptsize 4.15}} & \textbf{52.37 ± {\scriptsize 11.08}} & \textbf{60.97 ± {\scriptsize 9.15}}    \\
			\midrule
			DS-AL & 44.26 ± {\footnotesize 3.47} & 
            22.02 ± {\footnotesize 5.28} & 
            29.84 ± {\footnotesize 3.99} & 
            47.92 ± {\footnotesize 4.79} & 
            22.54 ± {\footnotesize 7.19} & 
            28.84 ± {\footnotesize 4.28} & 
            32.91 ± {\footnotesize 1.86} & 
            19.53 ± {\footnotesize 8.86} & 
            21.60 ± {\footnotesize 6.28} & 
            57.24 ± {\footnotesize 4.32} & 
            35.38 ± {\footnotesize 11.28} & 
            41.27 ± {\footnotesize 10.74} \\
			\midrule
			\rowcolor{LightCyan}DS-AL w/ \name & \textbf{44.62 ± {\footnotesize 6.73}} & \textbf{26.23 ± {\footnotesize 7.01}} & \textbf{31.15 ± {\footnotesize 5.23}} & \textbf{49.35 ± {\footnotesize 4.28}} & \textbf{26.47 ± {\footnotesize 6.39}} & \textbf{33.64 ± {\footnotesize 5.67}} & \textbf{34.23 ± {\footnotesize 1.22}} & \textbf{21.31 ± {\footnotesize 6.15}} & \textbf{26.85 ± {\footnotesize 3.28}} & \textbf{60.25 ± {\footnotesize 2.71}} & \textbf{40.28 ± {\footnotesize 13.43}} & \textbf{44.38 ± {\footnotesize 8.25}} \\
			\midrule
			TagFex & 43.62 ± {\footnotesize 3.09} & 
            22.76 ± {\footnotesize 4.09} & 
            28.58 ± {\footnotesize 7.28} & 
            49.81 ± {\footnotesize 4.64} & 
            26.13 ± {\footnotesize 7.30} & 
            33.97 ± {\footnotesize 5.12} & 
            35.41 ± {\footnotesize 1.28} & 
            26.23 ± {\footnotesize 4.33} & 
            30.10 ± {\footnotesize 4.75} & 
            56.83 ± {\footnotesize 2.74} & 
            33.21 ± {\footnotesize 13.24} & 
            39.44 ± {\footnotesize 11.37} \\
			\midrule
			\rowcolor{LightCyan}TagFex w/ \name & \textbf{45.98 ± {\footnotesize 2.71}} & \textbf{25.20 ± {\footnotesize 5.94}} & \textbf{30.86 ± {\footnotesize 5.62}} & \textbf{50.25 ± {\footnotesize 5.65}} & \textbf{28.31 ± {\footnotesize 7.37}} & \textbf{35.48 ± {\footnotesize 4.09}} & \textbf{37.48 ± {\footnotesize 0.79}} & \textbf{31.93 ± {\footnotesize 4.65}} & \textbf{34.87 ± {\footnotesize 2.85}} & \textbf{61.15 ± {\footnotesize 5.63}} & \textbf{37.44 ± {\footnotesize 13.24}} & \textbf{43.37 ± {\footnotesize 10.13}} \\
			\bottomrule
		\end{tabular}
	}
\end{table*}
\begin{figure*}[!t]
	\centering
	\begin{subfigure}{0.16\linewidth}
		\includegraphics[width=\linewidth]{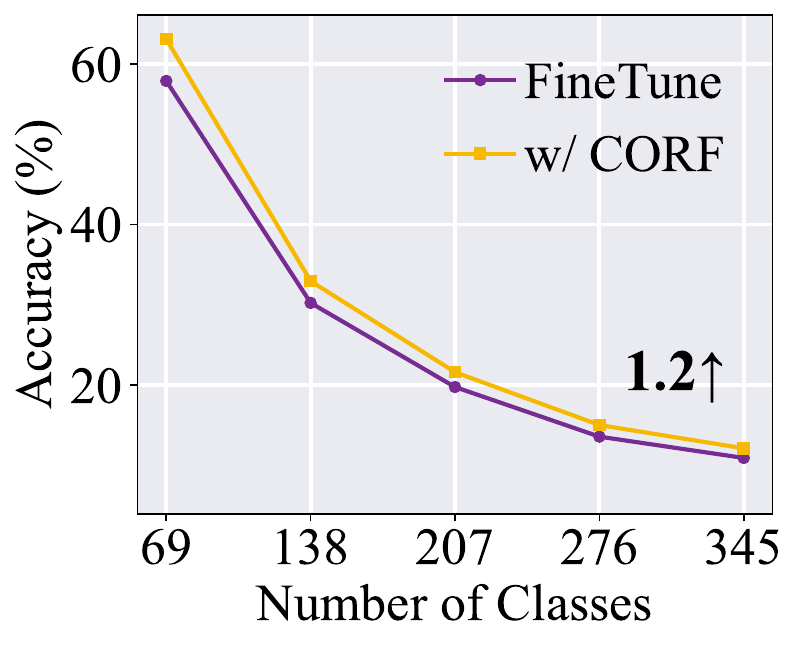}
		\caption{FineTune}
		\label{fig:finetune}
	\end{subfigure}
	\hfill
	\begin{subfigure}{0.16\linewidth}
		\includegraphics[width=\linewidth]{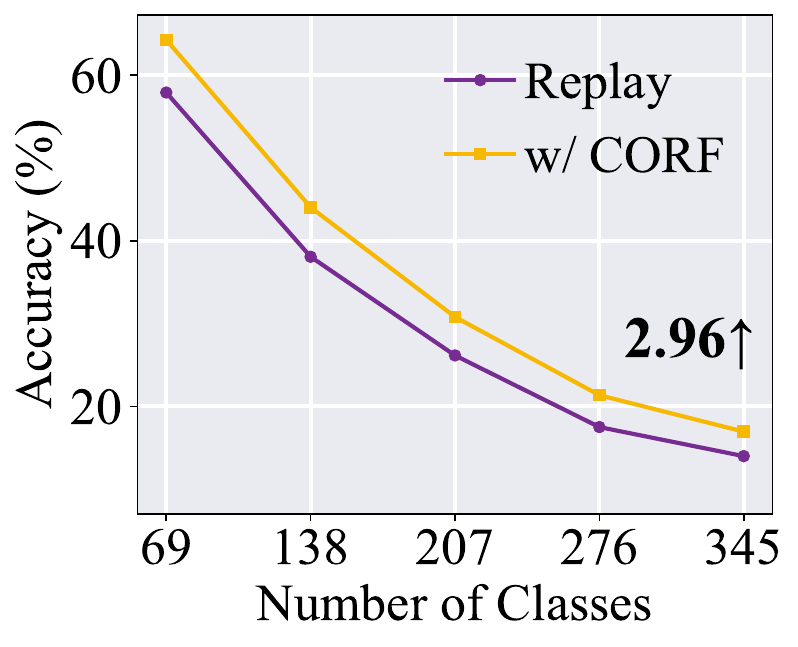}
		\caption{Replay}
		\label{fig:replay}
	\end{subfigure}
	\hfill
	\begin{subfigure}{0.16\linewidth}
		\includegraphics[width=\linewidth]{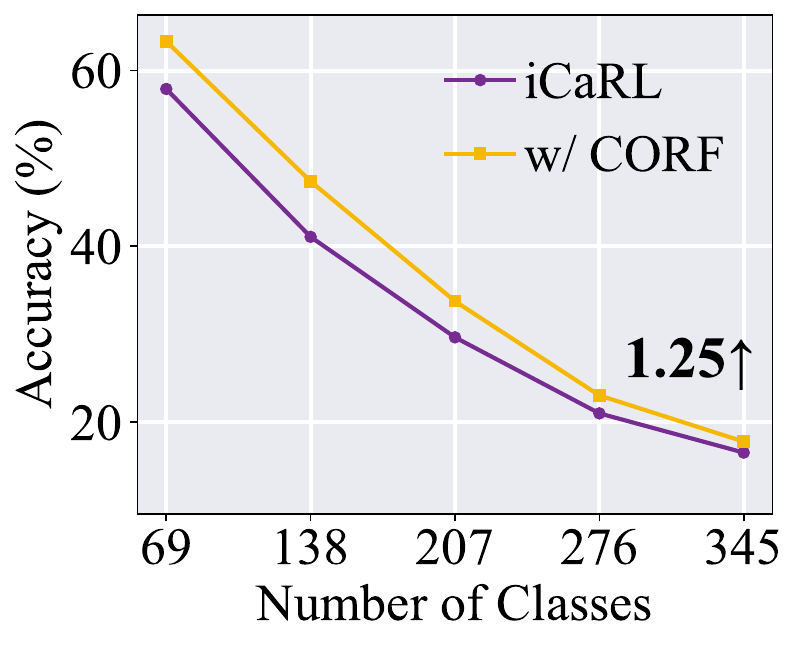}
		\caption{iCaRL}
		\label{fig:icarl}
	\end{subfigure}
	\hfill
	\begin{subfigure}{0.16\linewidth}
		\includegraphics[width=\linewidth]{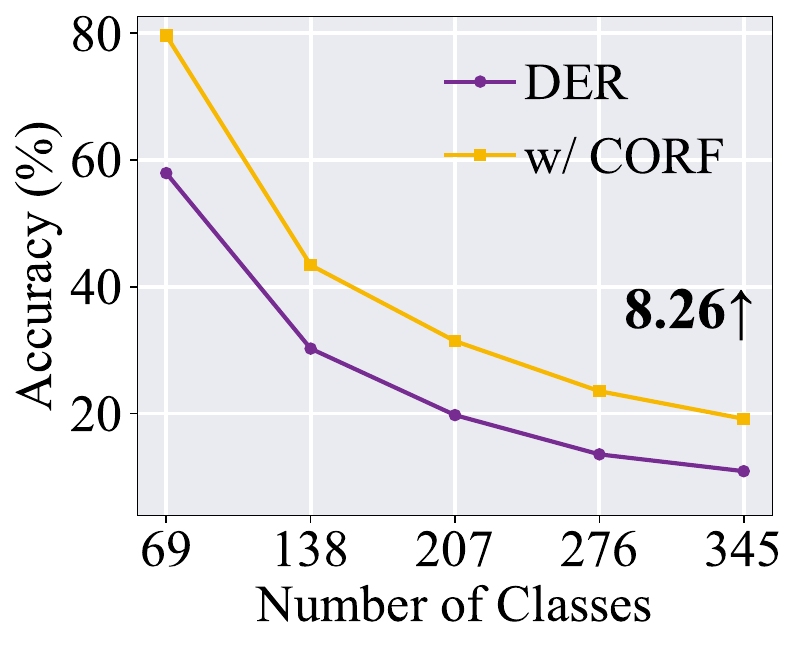}
		\caption{DER}
		\label{fig:der}
	\end{subfigure}
	\hfill
	\begin{subfigure}{0.16\linewidth}
		\includegraphics[width=\linewidth]{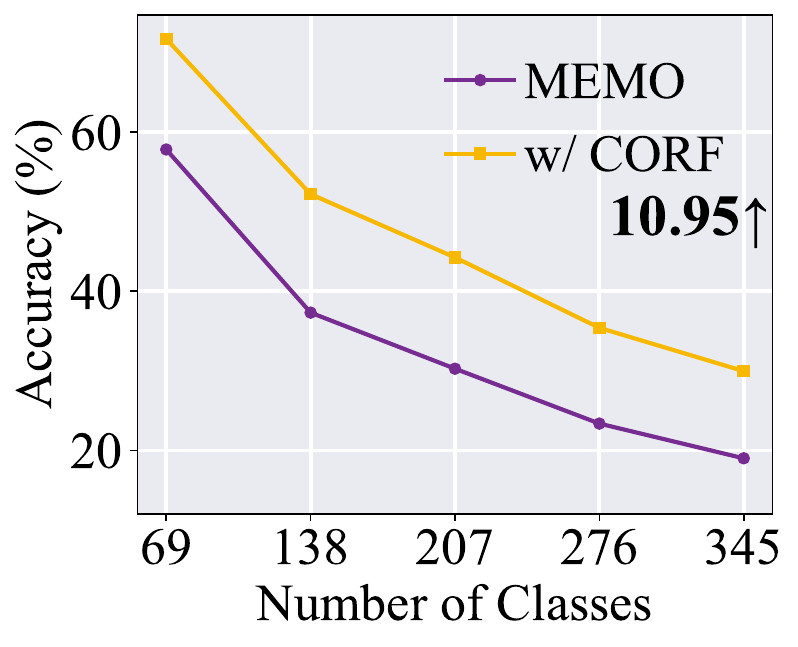}
		\caption{MEMO}
		\label{fig:memo}
	\end{subfigure}
	\hfill
	\begin{subfigure}{0.16\linewidth}
		\includegraphics[width=\linewidth]{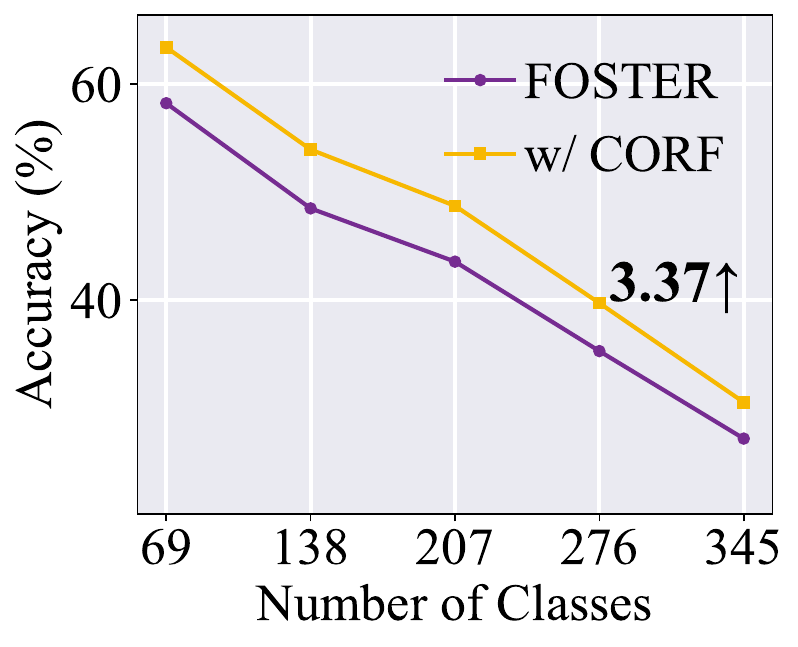}
		\caption{FOSTER}
		\label{fig:foster}
	\end{subfigure}
	\hfill
	\begin{subfigure}{0.16\linewidth}
		\includegraphics[width=\linewidth]{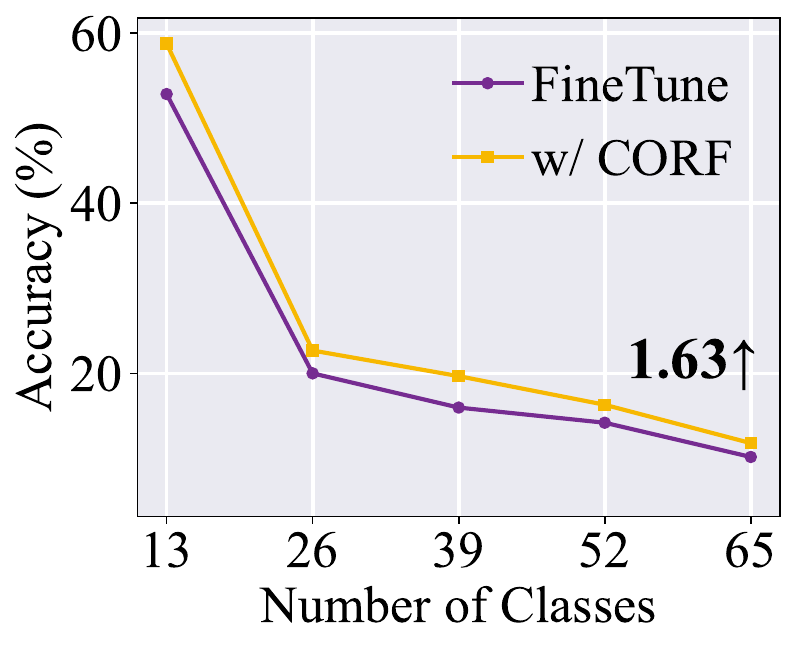}
		\caption{FineTune}
		\label{fig:finetune13}
	\end{subfigure}
	\hfill
	\begin{subfigure}{0.16\linewidth}
		\includegraphics[width=\linewidth]{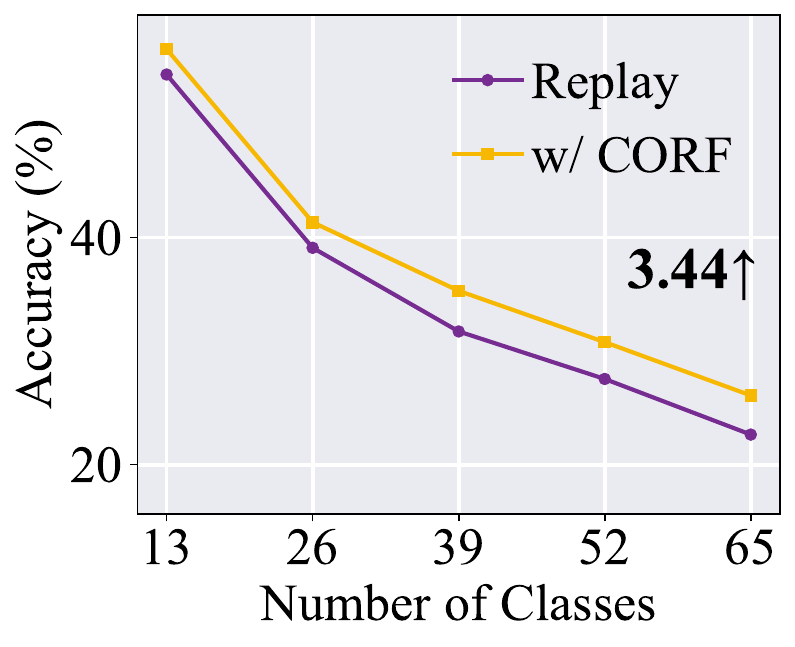}
		\caption{Replay}
		\label{fig:replay13}
	\end{subfigure}
	\hfill
	\begin{subfigure}{0.16\linewidth}
		\includegraphics[width=\linewidth]{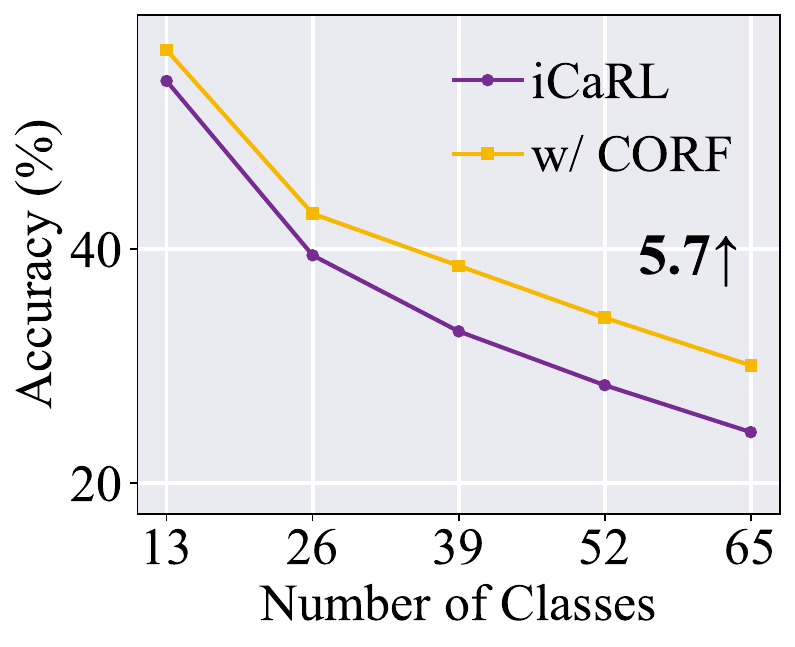}
		\caption{iCaRL}
		\label{fig:icarl13}
	\end{subfigure}
	\hfill
	\begin{subfigure}{0.16\linewidth}
		\includegraphics[width=\linewidth]{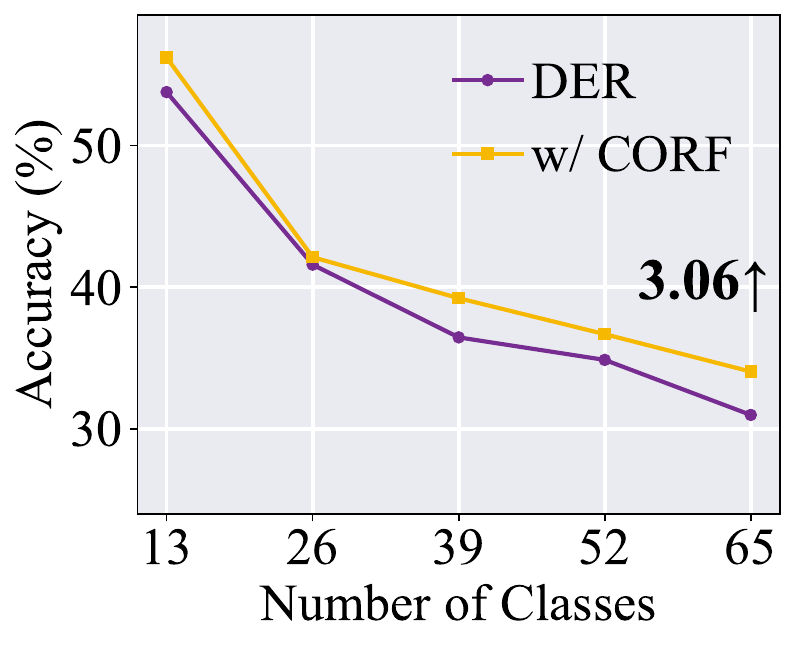}
		\caption{DER}
		\label{fig:der13}
	\end{subfigure}
	\hfill
	\begin{subfigure}{0.16\linewidth}
		\includegraphics[width=\linewidth]{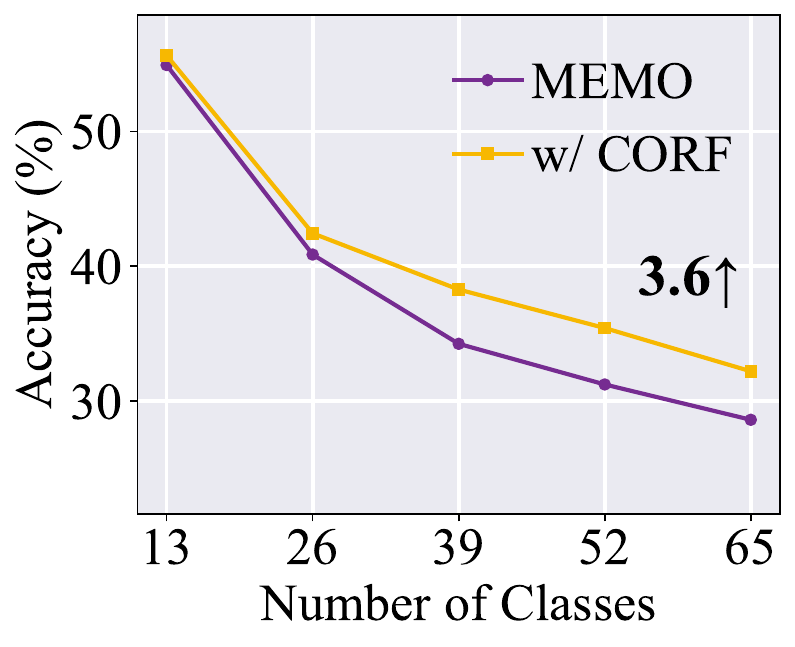}
		\caption{MEMO}
		\label{fig:memo13}
	\end{subfigure}
	\hfill
	\begin{subfigure}{0.16\linewidth}
		\includegraphics[width=\linewidth]{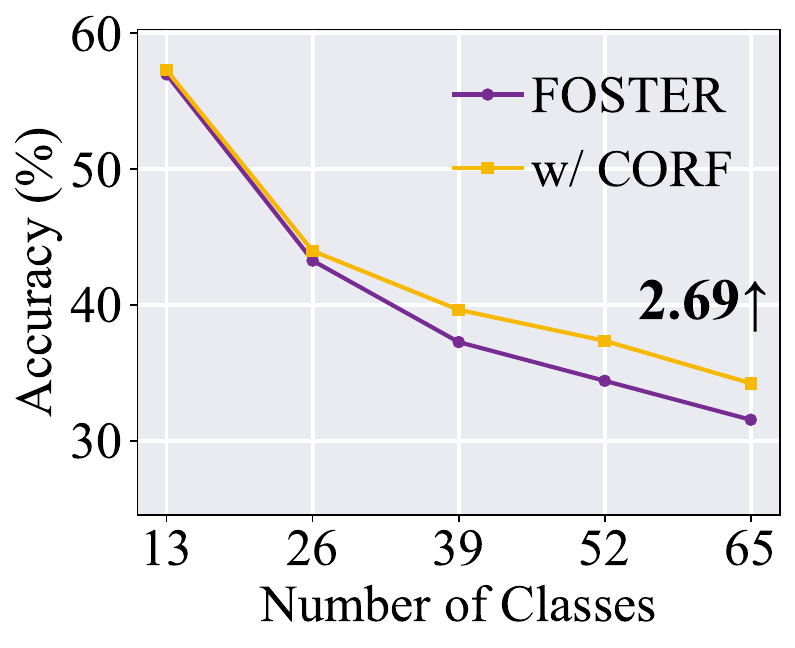}
		\caption{FOSTER}
		\label{fig:foster13}
	\end{subfigure}
	\hfill
	\begin{subfigure}{0.16\linewidth}
		\includegraphics[width=\linewidth]{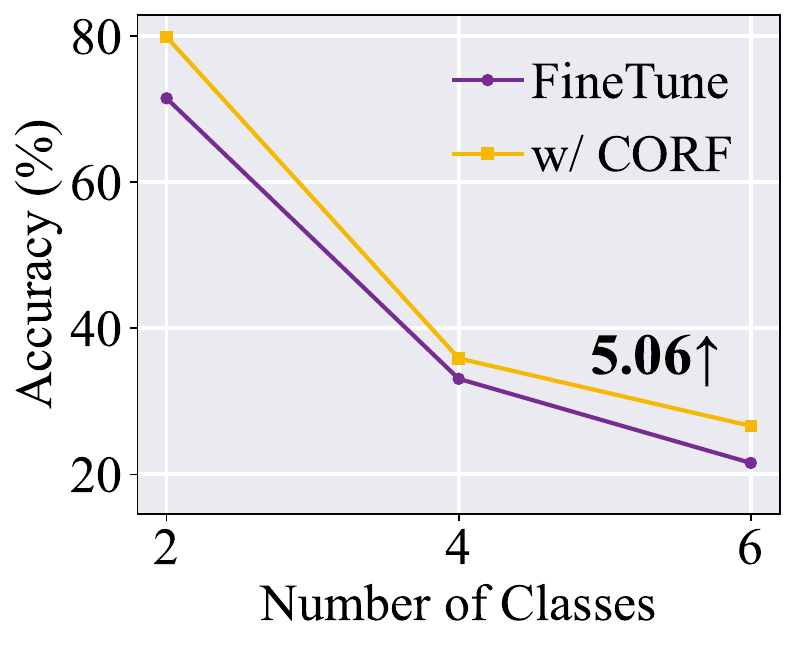}
		\caption{FineTune}
		\label{fig:finetune2}
	\end{subfigure}
	\hfill
	\begin{subfigure}{0.16\linewidth}
		\includegraphics[width=\linewidth]{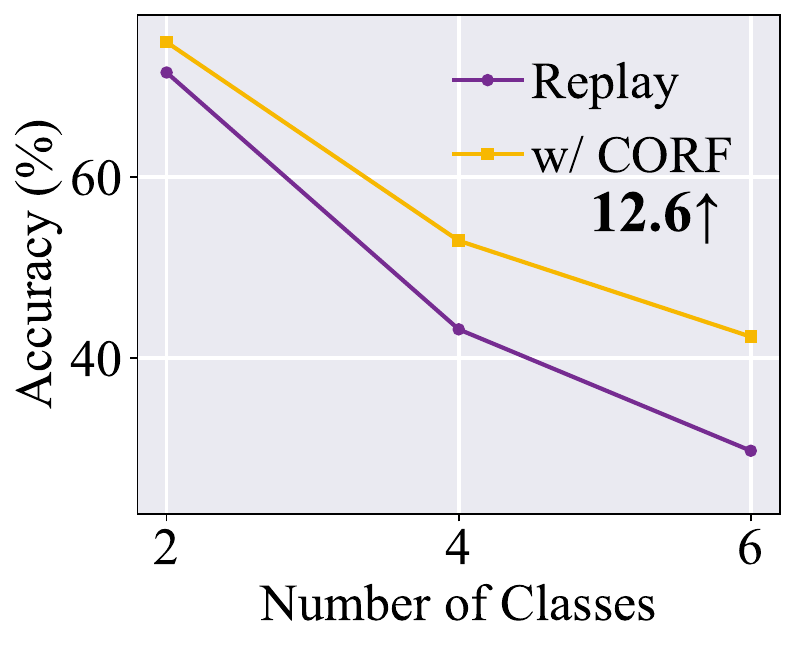}
		\caption{Replay}
		\label{fig:replay2}
	\end{subfigure}
	\hfill
	\begin{subfigure}{0.16\linewidth}
		\includegraphics[width=\linewidth]{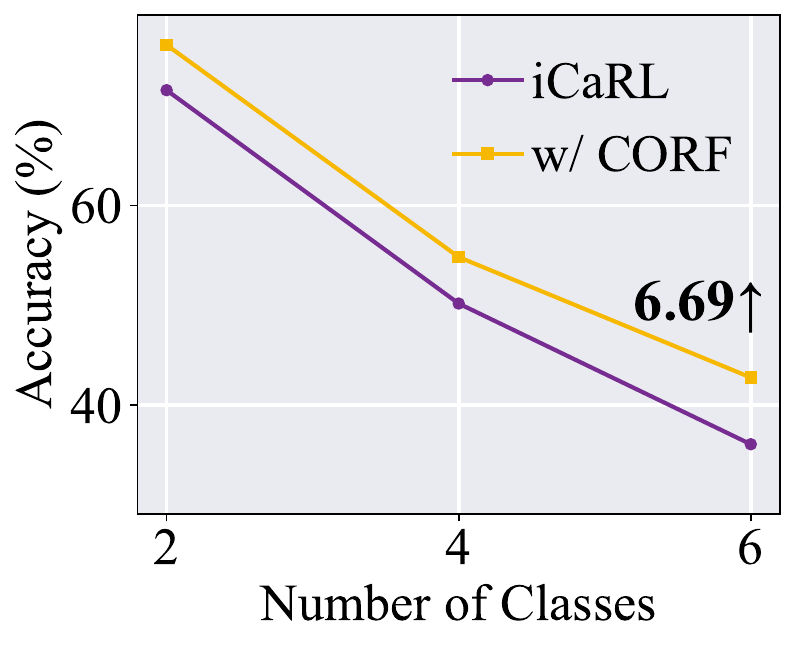}
		\caption{iCaRL}
		\label{fig:icarl2}
	\end{subfigure}
	\hfill
	\begin{subfigure}{0.16\linewidth}
		\includegraphics[width=\linewidth]{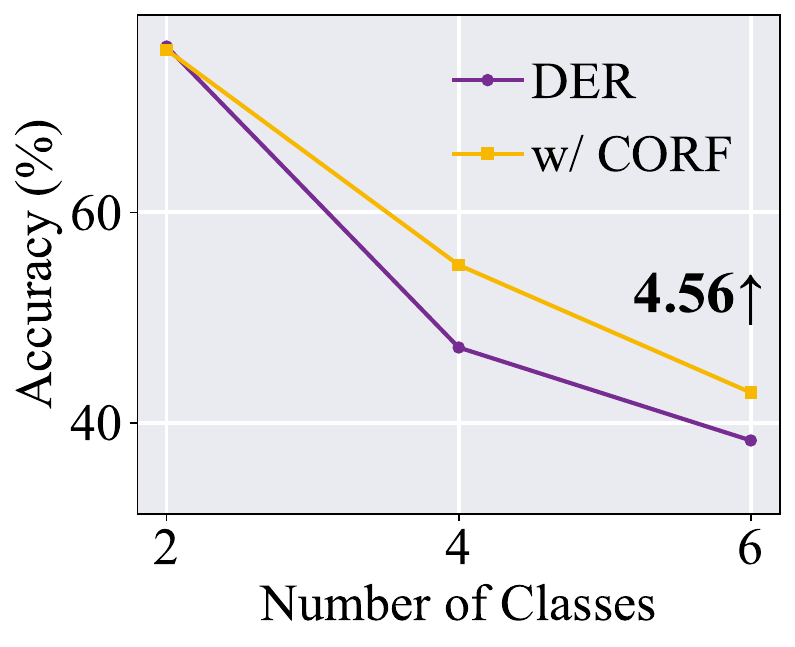}
		\caption{DER}
		\label{fig:der2}
	\end{subfigure}
	\hfill
	\begin{subfigure}{0.16\linewidth}
		\includegraphics[width=\linewidth]{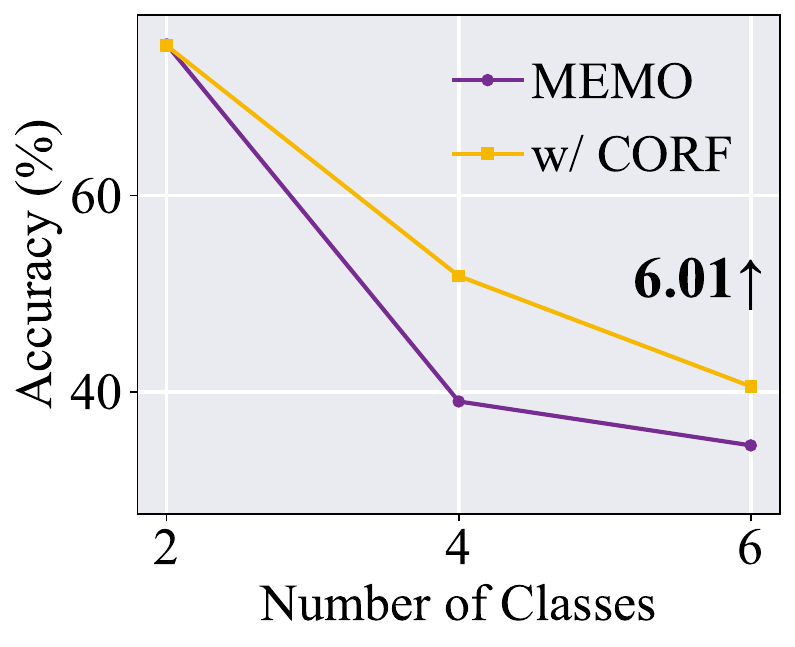}
		\caption{MEMO}
		\label{fig:memo2}
	\end{subfigure}
	\hfill
	\begin{subfigure}{0.16\linewidth}
		\includegraphics[width=\linewidth]{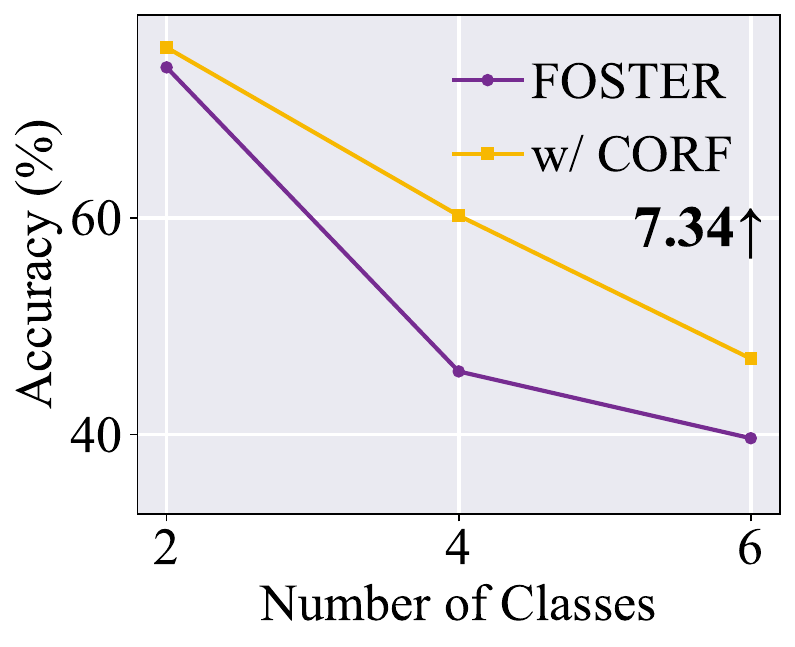}
		\caption{FOSTER}
		\label{fig:foster2}
	\end{subfigure}
\caption{
(a–f) Performance gains of C\ORF\ over baseline methods on DomainNet (5 tasks);
(g–l) results on OfficeHome (5 tasks);
(m–r) results on PACS (3 tasks).
We report the performance gap after the last incremental stage at the end of each curve.
}
	\label{fig:boost}
\end{figure*}

\paragraph{Comparison Methods}
To evaluate the effectiveness and versatility of C\ORF, we integrate it with a range of representative baselines, including FineTune, Replay, and iCaRL~\cite{DBLP:conf/cvpr/RebuffiKSL17}, as well as several state-of-the-art approaches including DER~\cite{DBLP:conf/cvpr/YanX021}, MEMO~\cite{DBLP:conf/iclr/0001WYZ23}, FOSTER~\cite{DBLP:conf/eccv/WangZYZ22}, DS-AL~\cite{zhuang2024ds} and TagFex~\cite{zheng2025task}. All methods are implemented with the same backbone network and training protocol for a fair comparison. 

\paragraph{Evaluation Protocol}
Following the benchmark established by \cite{DBLP:conf/cvpr/RebuffiKSL17}, we measure performance of different methods using the Top-$1$ accuracy after the $n$-th incremental stage, denoted as $A_n$. Based on this, we define $A_N$ as the accuracy after the final incremental stage and $\bar{A} = \tfrac{1}{N}\sum_{n=1}^N A_n$ as the average accuracy across all stages. Since CDCIL additionally involves domain generalization, we evaluate the model on both the seen domains (SD) and the unseen domain (UD) at the final stage, and compute their harmonic mean (HM) for balanced assessment. In practice, we adopt a domain-rotation protocol. As illustrated in Fig.~\ref{fig:EvaluationProtocol}, each domain is treated once as the unseen domain, while the remaining domains serve as seen domains for training. 
After training on all tasks, SD and UD accuracies are evaluated, and HM is computed accordingly. 
The reported results are averaged over all unseen-domain configurations, and the standard deviation across these rotations is also reported to reflect performance stability.

\subsection{Experimental Results}
\paragraph{Benchmark Comparison}
We evaluate the effectiveness of C\ORF\ by integrating it with several representative CIL baselines on three widely used benchmarks. As shown in Tab.~\ref{tab:benchmark}, C\ORF\ consistently improves performance across both classical methods and state-of-the-art approaches. Fig.~\ref{fig:boost} further presents the incremental learning trends on DomainNet, OfficeHome and PACS. Across all datasets, C\ORF\ delivers consistent improvements over the corresponding baselines throughout the incremental stages. The gains are particularly pronounced on structure-based methods such as DER, demonstrating its strong capability to construct a domain-agnostic and incrementally robust representation space under diverse domain shifts and task splits.

\begin{figure*}[t]
	\centering
	\begin{subfigure}{0.32\linewidth}
		\includegraphics[width=\linewidth]{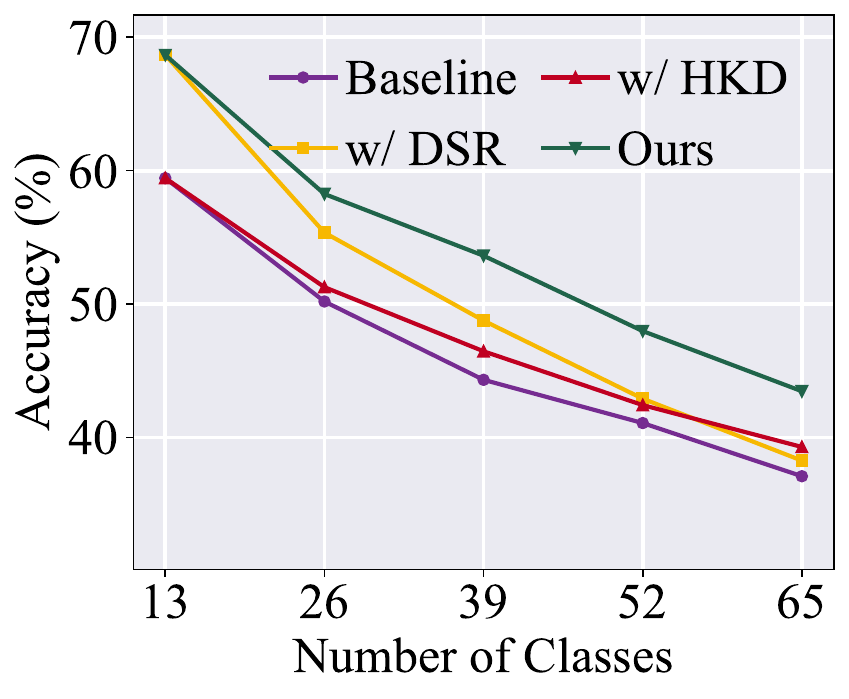}
		\caption{Seen domains}
		\label{fig:ablation_seen_domain}
	\end{subfigure}
	\hfill
	\begin{subfigure}{0.32\linewidth}
		\includegraphics[width=\linewidth]{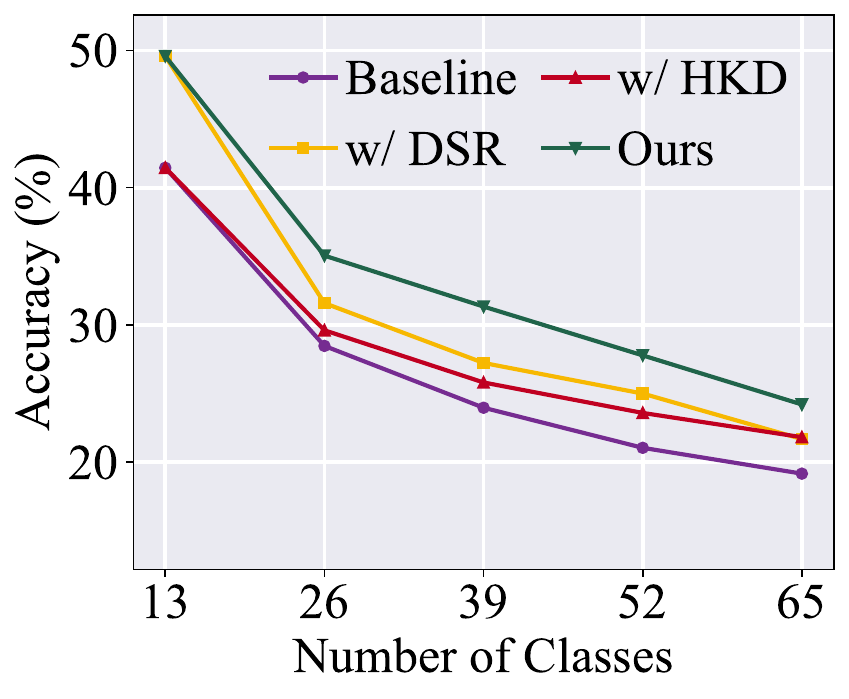}
		\caption{Unseen domain}
		\label{fig:ablation_unseen_domain}
	\end{subfigure}
	\hfill
	\begin{subfigure}{0.32\linewidth}
		\includegraphics[width=\linewidth]{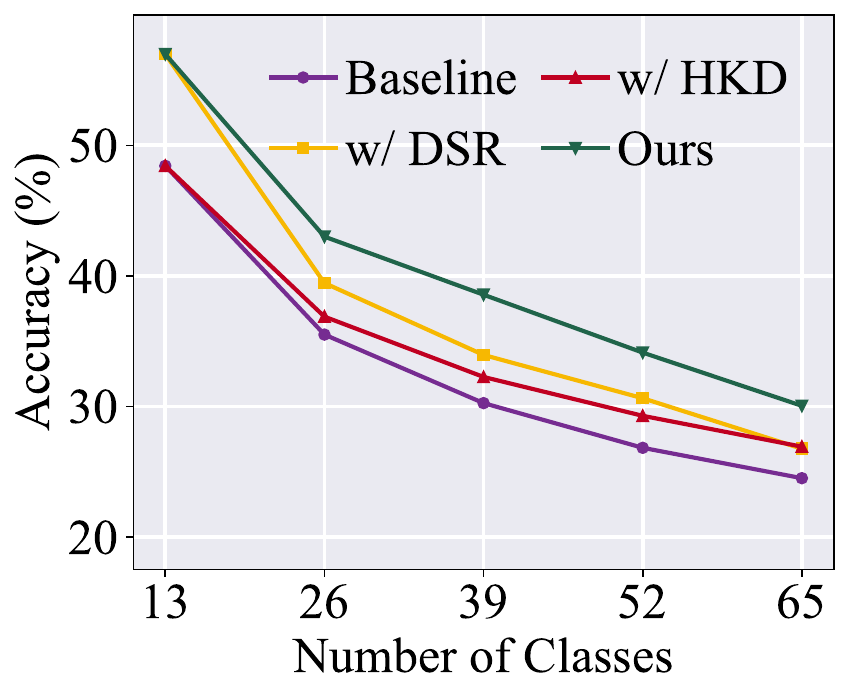}
		\caption{Harmonic mean}
		\label{fig:ablation_harmonic}
	\end{subfigure}
	\caption{
    (\subref{fig:ablation_seen_domain})–(\subref{fig:ablation_harmonic}) present the ablation study results on seen domains, unseen domains, and their harmonic mean, respectively. Every component of C\ORF\ improves the performance.
    }
	\label{fig:ablation_full}
\end{figure*}

\paragraph{Ablation Study}
In this section, we conduct an ablation study on OfficeHome 5 tasks by incrementally adding each component to evaluate its effectiveness within C\ORF. As depicted in Fig.~\ref{fig:ablation_full} (\subref{fig:ablation_seen_domain}–\subref{fig:ablation_harmonic}), we first reproduce the standard iCaRL method as \texttt{Baseline} and then progressively incorporate the individual designs. Incorporating DSR yields consistent gains on both SD and UD, demonstrating its ability to improve cross-domain generalization. When HKD is applied alone, the accuracy degradation is clearly alleviated, particularly on UD, showing its strength in mitigating forgetting. When combined, DSR and HKD further complement each other, leading to an additional boost in overall performance and a substantial improvement in HM. In general, the optimization of each component, such as DSR and HKD, delivers clear benefits and all contribute to the strong performance gain of C\ORF.

\begin{figure*}
	\centering
 \begin{subfigure}{0.33\linewidth}
		\includegraphics[width=\linewidth]{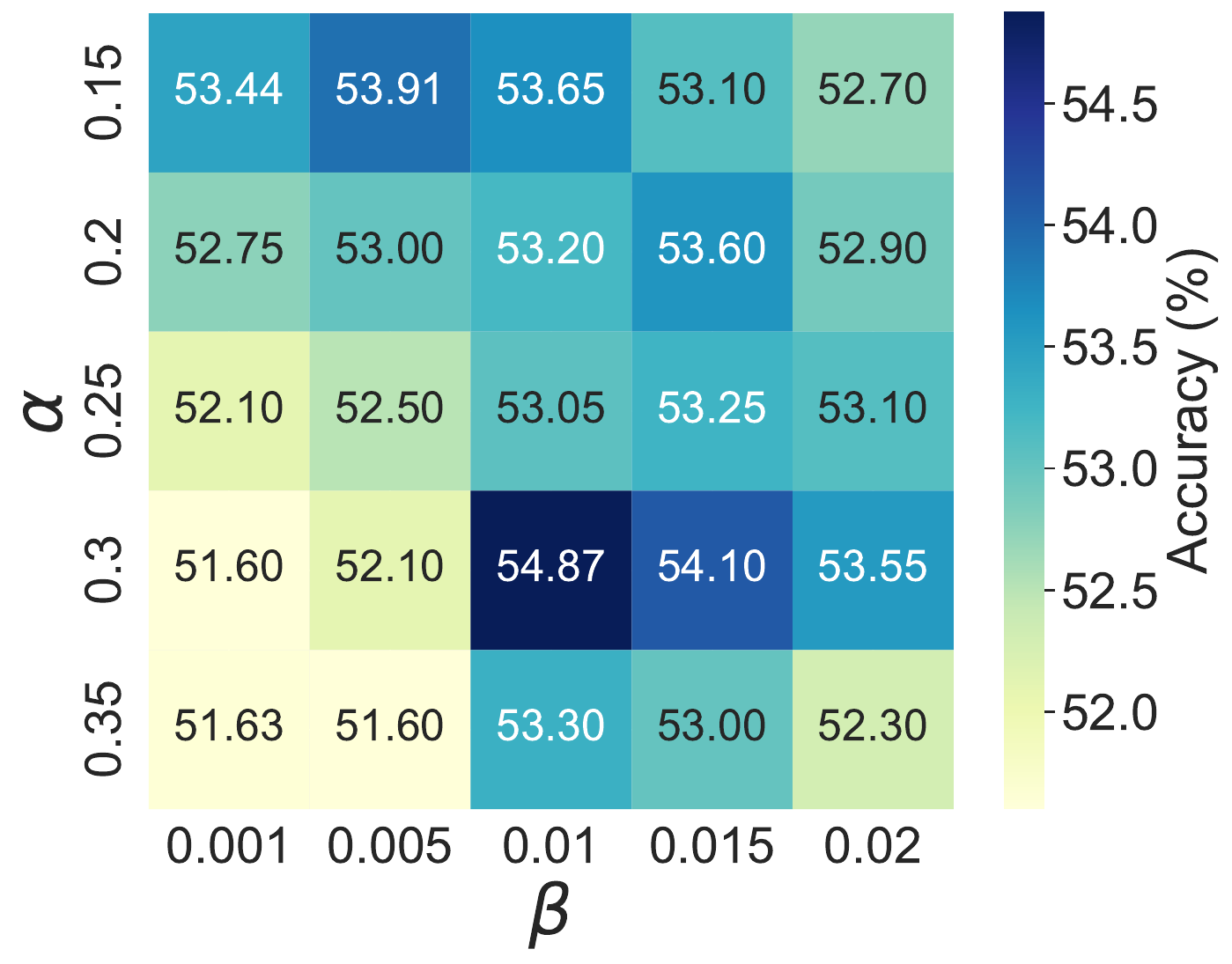}
		\caption{Sensitivity analysis}
		\label{fig:hyperparameter_sensitivity}
	\end{subfigure}
 \hfill
	\begin{subfigure}{0.32\linewidth}
		\includegraphics[width=1\columnwidth]{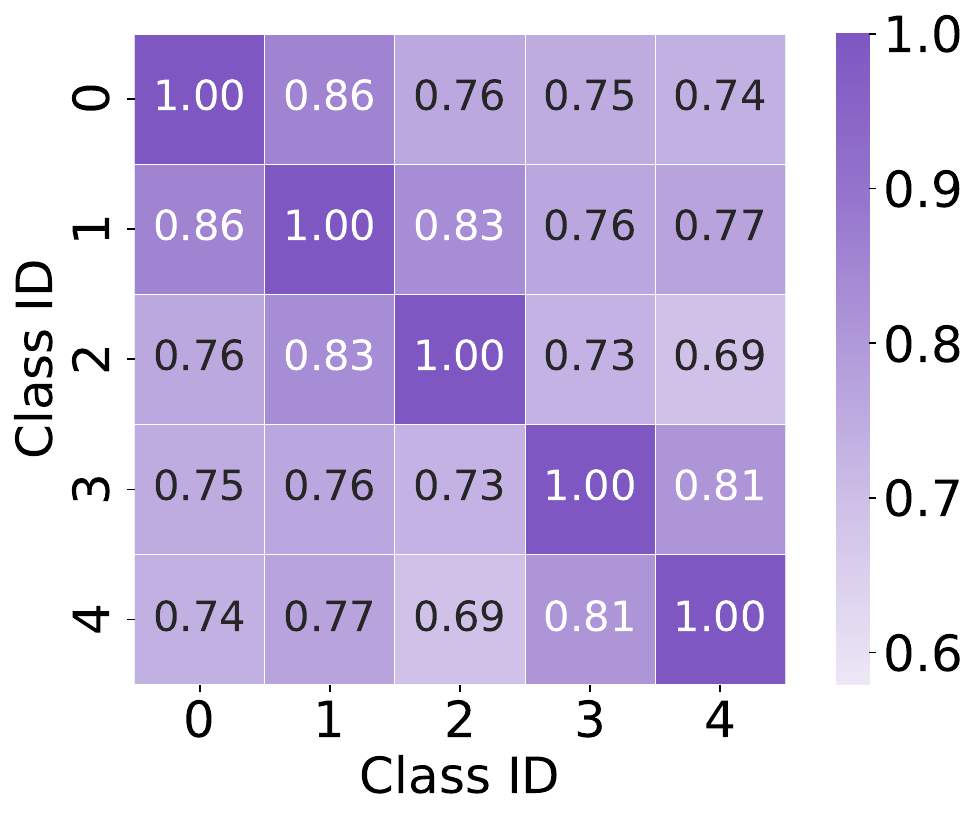}
		\caption{Baseline}
		\label{fig:Baseline similarity matrix}
	\end{subfigure}
	\hfill
	\begin{subfigure}{0.32\linewidth}
		\includegraphics[width=1\linewidth]{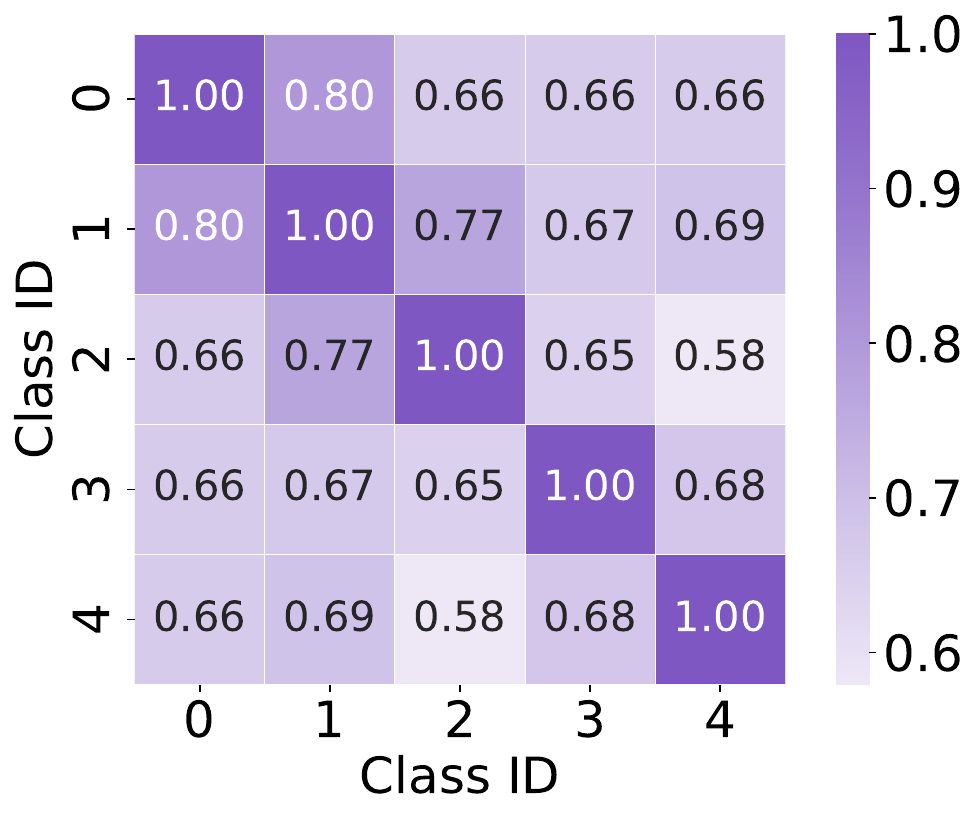}
		\caption{Ours}
		\label{fig:Our similarity matrix}
	\end{subfigure}
	\caption{
 (\subref{fig:hyperparameter_sensitivity}) shows the hyperparameter sensitivity analysis over the representation refinement weight $\alpha$ and the HKD coefficient $\beta$, while (\subref{fig:Baseline similarity matrix})-(\subref{fig:Our similarity matrix}) show similarity matrices of five randomly selected classes under the baseline and ours, respectively.
    }
	\label{fig:similarity}
\end{figure*}

\paragraph{Hyperparameter Sensitivity Analysis}
We conduct a sensitivity analysis of the two key hyperparameters on PACS: the representation refinement weight $\alpha$ and the HKD coefficient $\beta$. As shown in Fig.~\ref{fig:hyperparameter_sensitivity}, the overall HM remains relatively stable across a wide range of values, demonstrating the strong robustness of C\ORF\ to hyperparameter variations. The best performance is observed at $\alpha= 0.3$ and $\beta = 0.01$, but nearby configurations yield comparably strong results, suggesting that C\ORF\ does not require complex or extensive tuning to achieve competitive performance, making it more convenient and efficient for real-world applications.

\begin{figure}[t]
	\centering
	\begin{subfigure}{0.48\linewidth}
		\includegraphics[width=1\linewidth]{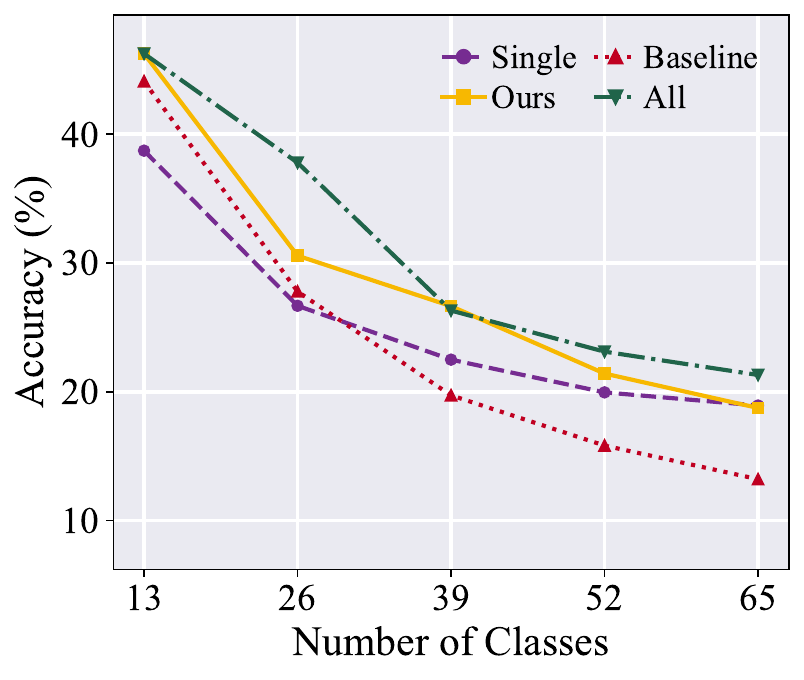}
        \caption{Upper bound exploration}
        \label{fig:dominant}
	\end{subfigure}
	\hfill
	\begin{subfigure}{0.48\linewidth}
		\includegraphics[width=1\linewidth]{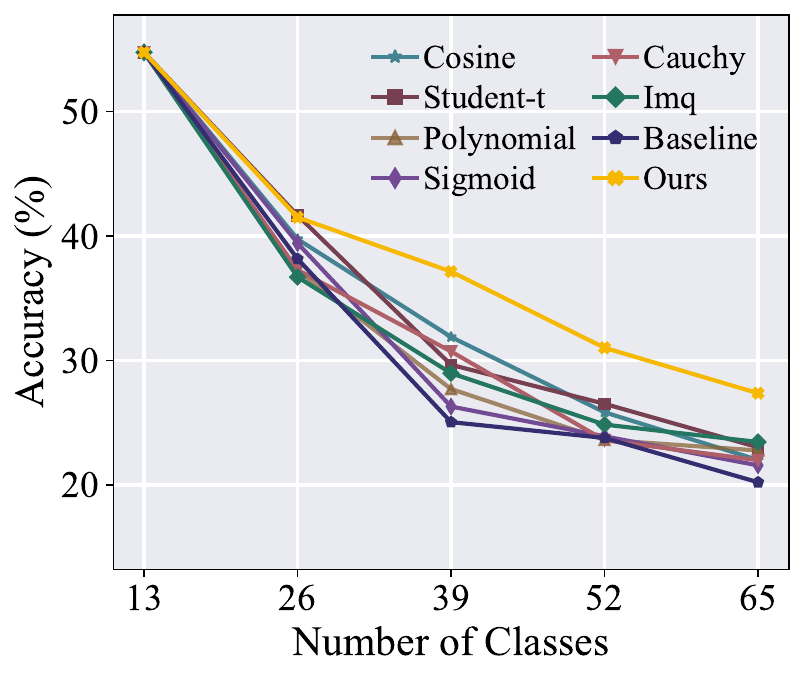}
        \caption{Kernel comparison}
        \label{fig:kernel}
	\end{subfigure}
	\caption{
(\subref{fig:dominant}) presents the Harmonic mean under different protocols.  (\subref{fig:kernel}) shows the impact of kernel selection on performance.
}
	\label{fig:Selection}
    \vspace{-5mm}
\end{figure}

\paragraph{Feature Similarity Analysis}
To further examine the impact of our method on feature representation, we visualize the pairwise similarity matrices of five randomly selected classes across five incremental stages on OfficeHome 5 tasks. As shown in Fig.~\ref{fig:similarity} (\subref{fig:Baseline similarity matrix}-\subref{fig:Our similarity matrix}) , both matrices are computed using the same set of classes to ensure a fair comparison between the iCaRL-based baseline and C\ORF. Compared to \texttt{Baseline} (Fig.~\ref{fig:Baseline similarity matrix}), C\ORF\ (Fig.~\ref{fig:Our similarity matrix}) yields significantly lower inter-class similarity, indicating more compact and better-separated representations. This demonstrates the effectiveness of DSR in enhancing class discriminability and mitigating feature entanglement across domains and tasks.

\paragraph{Upper Bound Exploration}
To assess the potential upper bound of our method, we conduct experiments under four settings on OfficeHome 5 tasks, using iCaRL as the baseline and selecting `Art' as the unseen domain. As shown in Fig.~\ref{fig:dominant}, both \texttt{Ours} and \texttt{Baseline} follow the standard CDCIL protocol. \texttt{Single} denotes continual learning within `Art' only, while \texttt{All} represents learning across all domains jointly without any generalization constraint. We report HM over incremental stages. C\ORF\ consistently outperforms the \texttt{Baseline} at all stages, and in some cases even surpasses the \texttt{Single}, demonstrating its ability to generalize to unseen domains by leveraging knowledge from other domains. Although there remains a gap compared to the \texttt{All}, C\ORF\ significantly narrows the performance gap relative to \texttt{Baseline}.

\paragraph{Kernel Selection}
We evaluate the impact of kernel choice on CDCIL performance using OfficeHome 5 tasks. As shown in Fig.~\ref{fig:kernel}, we compare six kernels: Cosine, Student-$t$, Polynomial, Sigmoid, Cauchy, and IMQ. We report HM across domains and incremental stages, among which Cosine and Student-$t$ kernels perform best, reflecting their complementary strengths: Cosine captures angular similarity and preserves global semantic orientation that is relatively insensitive to feature magnitude variations, while Student-$t$ focuses on local neighborhood structure with a heavy-tailed decay that is robust to outliers and abrupt representation shifts. 

In CDCIL, incremental updates and domain shifts can distort pairwise distances unevenly: global semantic directions may remain stable while local neighborhoods drift, or vice versa. 
By combining Cosine and Student-$t$ kernels, our hybrid HKD design jointly preserves global relational geometry and robust local topology, achieving clear performance gains and more balanced generalization across incremental stages.

\paragraph{Resource Consumption}
On average, C\ORF\ requires about 1.2$\times$ the training time of a standard CIL baseline such as DER, mainly due to (i) Grad-CAM-based contribution maps, (ii) region-level fusion in DSR, and (iii) relational alignment in HKD. Grad-CAM is computed only for a selected subset of samples within each mini-batch and implemented via a single backward pass on the last convolutional block, so the overhead scales with the selection ratio rather than the batch size.
C\ORF\ introduces no additional trainable parameters and does not require extra feature storage beyond standard backpropagation. In practice, peak GPU memory remains comparable to baselines. For example, under ResNet-18 on PACS, memory increases from 13.3\,GB (DER) to 16.6\,GB (DER w/ C\ORF). On DomainNet (`Clipart' as UD), FineTune with C\ORF\ takes 198 hours versus 167 hours for FineTune on a single RTX 3090 GPU. Increasing batch size from 32 to 64 keeps the relative overhead stable (1.18$\times$–1.23$\times$), indicating good scalability.

\begin{figure*}[!t]
	\centering
	\begin{subfigure}{0.32\linewidth}
		\includegraphics[width=1\columnwidth]{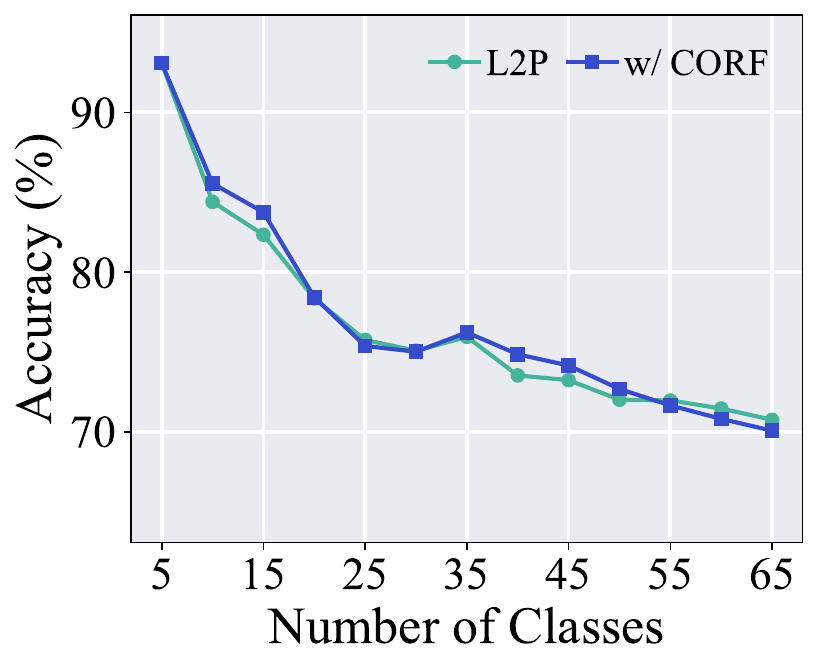}
        \caption{SD}
        \label{fig:arc_in_domain_l2p}
	\end{subfigure}
	\hfill
	\begin{subfigure}{0.32\linewidth}
		\includegraphics[width=1\columnwidth]{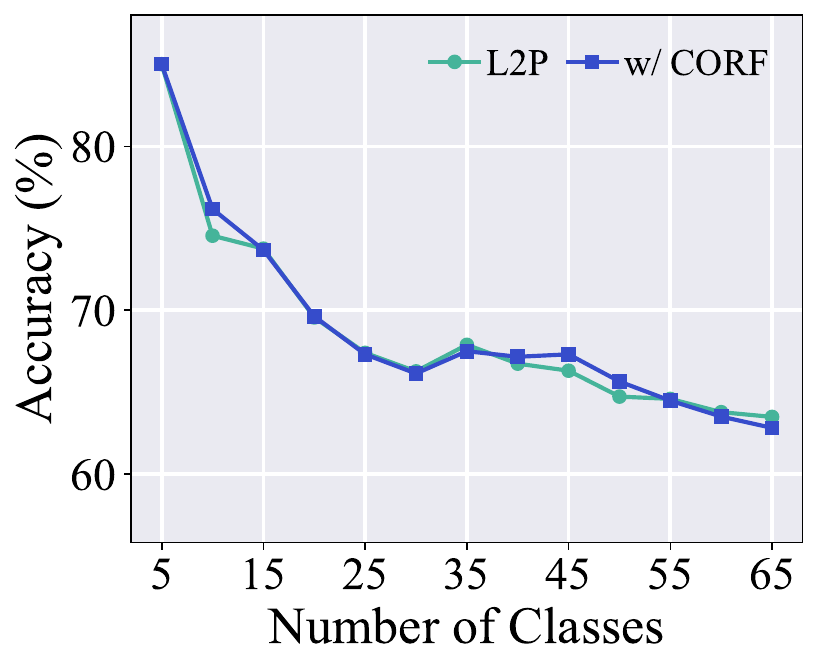}
        \caption{UD}
        \label{fig:arc_out_domain_l2p}
	\end{subfigure}
	\hfill
	\begin{subfigure}{0.32\linewidth}
		\includegraphics[width=1\columnwidth]{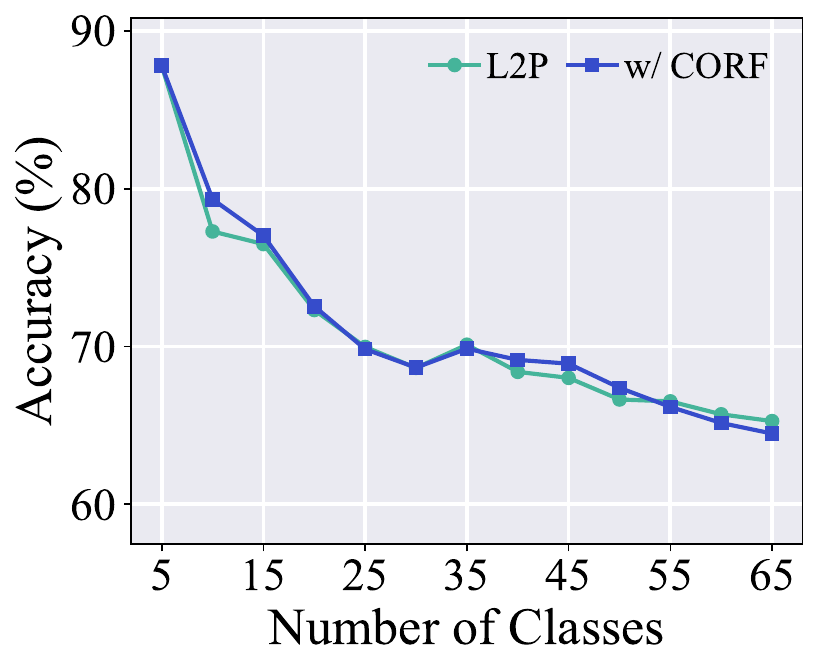}
        \caption{HM}
        \label{fig:arc_harmonic_l2p}
	\end{subfigure}
	\hfill
	\begin{subfigure}{0.32\linewidth}
		\includegraphics[width=1\columnwidth]{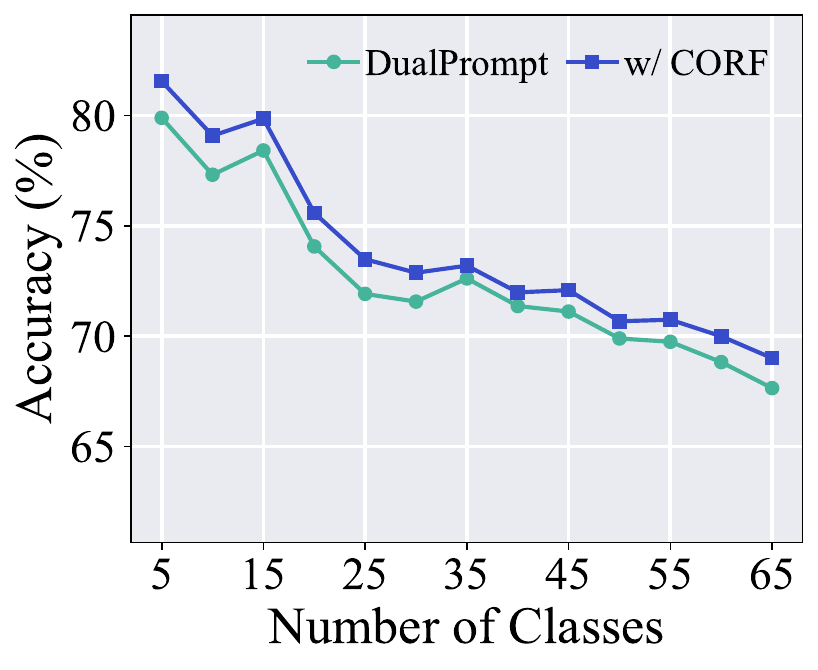}
        \caption{SD}
        \label{fig:arc_in_domain_dual}
	\end{subfigure}
	\hfill
	\begin{subfigure}{0.32\linewidth}
		\includegraphics[width=1\columnwidth]{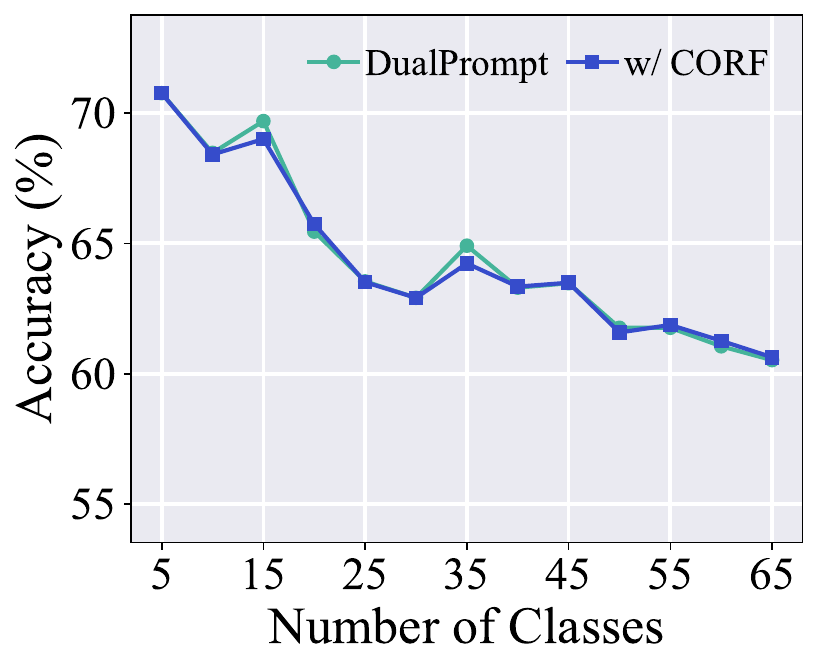}
        \caption{UD}
        \label{fig:arc_out_domain_dual}
	\end{subfigure}
	\hfill
	\begin{subfigure}{0.32\linewidth}
		\includegraphics[width=1\columnwidth]{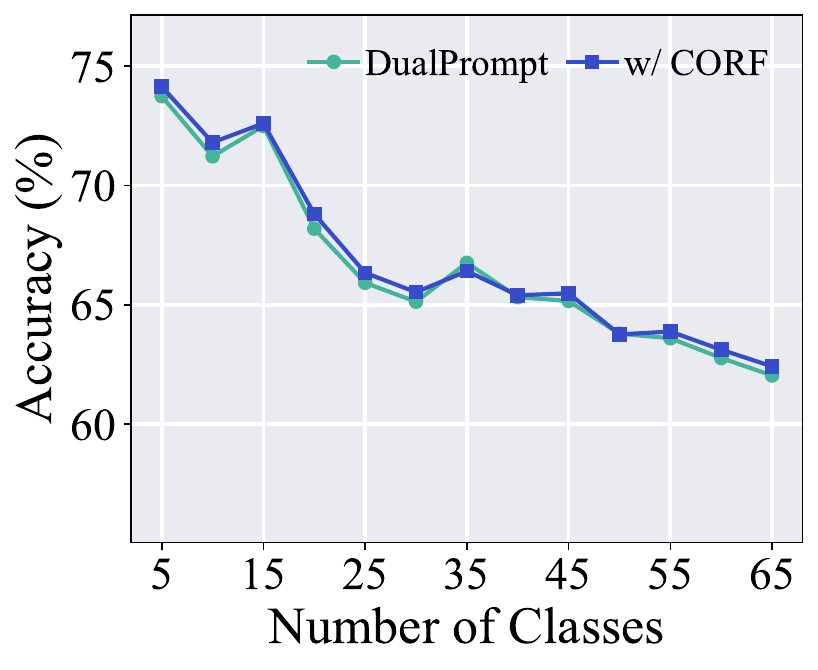}
        \caption{HM}
        \label{fig:arc_harmonic_dual}
	\end{subfigure}
	\caption{Compatibility of C\ORF\ with transformer-based CIL frameworks using ViT-B/16-IN1K as the backbone. 
(a--c) Results on L2P. 
(d--f) Results on DualPrompt. 
We report SD, UD, and HM across incremental stages.}
	\label{fig:arc}
\end{figure*}

\paragraph{Compatibility with Transformer Backbones}
To examine the applicability of C\ORF\ beyond CNN-based incremental learning, we further integrate it into two representative prompt-based ViT frameworks: L2P~\cite{DBLP:conf/cvpr/0002ZL0SRSPDP22} and DualPrompt~\cite{DBLP:conf/eccv/0002ZESZLRSPDP22}. Both methods are evaluated with ViT-B/16-IN1K as the backbone. As shown in Fig.~\ref{fig:arc}, we report the performance on seen domains (SD), unseen domains (UD), and their harmonic mean (HM) across incremental stages.

Overall, C\ORF\ can be seamlessly incorporated into both L2P and DualPrompt without causing performance collapse or disrupting the prompt-based learning pipeline. For L2P, C\ORF\ achieves comparable or slightly better performance than the baseline in most stages, especially in the early incremental phases on UD and HM, indicating that the proposed refinement and distillation mechanisms can provide additional robustness under domain shift. As the number of classes increases, the performance gap gradually becomes smaller, suggesting that the benefit of C\ORF\ is more pronounced when the incremental classification problem is relatively less saturated.

For DualPrompt, C\ORF\ also maintains stable performance and generally follows the baseline trend across SD, UD, and HM. The improvements are moderate but consistent in several stages, showing that C\ORF\ is compatible with prompt-based continual learning even when the trainable parameters are limited. Compared with CNN-based backbones, however, the gains on ViT-based prompt-tuning frameworks are less significant. This is likely because most backbone parameters are frozen, and the representation space is mainly adjusted through a small number of prompt parameters, which limits the extent to which DSR and HKD can reshape intermediate representations.

These results suggest that C\ORF\ is not restricted to convolutional architectures and can be applied to transformer-based CIL methods in a plug-and-play manner. Nevertheless, its effectiveness depends on the update paradigm of the backbone: when only lightweight prompts are optimized, C\ORF\ provides stable and sometimes beneficial regularization, but the improvement is naturally more constrained than in fully trainable CNN settings.

\section{Conclusion}
This paper addresses the challenge of CDCIL, which requires models to continuously learn new classes while generalizing to unseen domains and mitigating forgetting. We propose C\ORF, a unified framework that effectively enhances both generalization and forgetting resistance, achieving significant performance gains over existing baselines. In future work, we plan to extend C\ORF\ to more diverse incremental learning scenarios and more complex domain shifts, further improving its robustness and adaptability.

\setcounter{section}{0}
\renewcommand{\thesection}{\Roman{section}}
\makeatletter

\bibliographystyle{IEEEtran}

\bibliography{reference}

\newpage

\end{document}